\documentclass{article}

\usepackage{arxiv}
\usepackage[letterpaper]{geometry}
\usepackage{graphicx}
\usepackage{amsmath}
\usepackage{amssymb}
\usepackage{bm}
\usepackage{amsfonts}
\usepackage{subcaption}
\usepackage{algorithm}
\usepackage{algpseudocode}
\usepackage{mathtools}
\graphicspath{{{./}}}
\usepackage[normalem]{ulem}
\newif\ifclean
\cleantrue  
\ifclean
\newcommand{\COMMENT}[1]{}
\newcommand{\TODO}[1]{}
\newcommand{\REVISE}[1]{}
\newcommand{\QUESTION}[1]{}
\else
\usepackage{color}
\newcommand{\COMMENT}[1]{\textcolor{magenta}{{[ \sc{#1} ]}}} 
\newcommand{\QUESTION}[1]{\textcolor{cyan}{{[ {\bf Q:} { \bf{#1} }]}}} 
\newcommand{\REVISE}[1]{\textcolor{cyan}{{{#1}}}} 

\newcommand{\TODO}[1]{\textcolor{magenta}{{{\bf TODO:{#1}}}}} 

\newcommand{\red}[1]{\textcolor{red}{{#1}}}
\newcommand{\caution}{\red{\bf Draft: \today. Do not distribute.}}
\fi

\newcommand{\Aref}[1]{Alg.\,\ref{#1}}
\newcommand{\aref}[1]{App.\,\ref{#1}}
\newcommand{\fref}[1]{Fig.\,\ref{#1}}

\newcommand{\eref}[1]{Eq.\,(\ref{#1})}

\newcommand{\sref}[1]{Sec.\!~\ref{#1}}

\newcommand{\cref}[1]{Ref.\,\cite{#1}}
\newcommand{\crefs}[1]{Refs.\,\cite{#1}}

\newcommand{\Eb}{\mathbf{E}}

\newcommand{\Ib}{\mathbf{I}}

\newcommand{\Nc}{\mathcal{N}}

\newcommand{\Wc}{\mathcal{W}}

\newcommand{\fs}{\mathsf{f}}
\newcommand{\gs}{\mathsf{g}}

\newcommand{\ys}{\mathsf{y}}

\newcommand{\xs}{\mathsf{x}}

\newcommand{\Bs}{\mathsf{B}}

\newcommand{\Is}{\mathsf{I}}

\newcommand{\Sigmab}{{\boldsymbol{\Sigma}}}

\newcommand{\Gammab}{{\boldsymbol{\Gamma}}}
\newcommand{\etab}{{\boldsymbol{\eta}}}
\newcommand{\tr}{{\operatorname{tr}}}

\newcommand{\partialb}{{{\boldsymbol{\partial}}}}
\newcommand{\NN}{{\mathsf{N}\!\mathsf{N}}}

\newcommand{\invariants}{{\mathcal{I}}}
\newcommand{\RE}{{R}}
\newcommand{\Exp}{\mathbb{E}}
\newcommand{\Var}{\mathbb{V}}
\newcommand{\Cov}{\mathbb{C}}

\newcommand{\stress}{\mathbf{S}}
\newcommand{\strain}{\mathbf{E}}

\newcommand{\hiddenstate}{\mathsf{h}}
\newcommand{\energy}{\Psi}
\newcommand{\weights}{\mathsf{w}}
\newcommand{\parameters}{\boldsymbol{\theta}}
\newcommand{\data}{\mathsf{D}}
\newcommand{\dd}{\mathrm{d}}
\newcommand{\dt}{\mathrm{d}t}
\newcommand{\prob}{\pi}
\newcommand{\drift}{\mathsf{f}}
\newcommand{\diffusion}{\gamma}
\newcommand{\model}{\mathsf{m}}
\newcommand{\KLD}{D_\text{KL}}
\newcommand{\KL}{D_\text{KL}}
\newcommand{\loss}{\mathcal{L}}
\newcommand{\posteriortag}{*}
\newcommand{\priortag}{0}
\newcommand{\inputs}{\mathsf{X}}
\newcommand{\outputs}{\mathsf{Y}}
\newcommand{\inpt}{\mathsf{x}}
\newcommand{\outpt}{\mathsf{y}}
\newcommand{\rhs}{\mathsf{R}}
\newcommand{\observation}{\mathsf{Z}}
\newcommand{\nrepl}{N_R}

\newcommand{\surrogateposterior}{\mathsf{q}}
\newcommand{\PE}{U}
\newcommand{\CDF}{\text{CDF}}

\newcommand{\grad}{\boldsymbol{\nabla}}

\newcommand{\wH}{\weights_{\rhs}}
\newcommand{\wHj}{\weights_{\rhs,j}}
\newcommand{\wO}{\weights_{\observation}}
\newcommand{\wOj}{\weights_{\observation,j}}
\newcommand{\ybr}{\bar{y}}

\newcommand{\Schl}{Schl\"{o}gl}

\renewcommand{\shorttitle}{UQ of NN models via Langevin sampling}
\title{\bf Uncertainty quantification of neural network models of evolving processes via Langevin sampling}

\author{
Cosmin Safta\thanks{Equal contribution} \\
{\it Sandia National Laboratories},\\
Livermore, CA 94551
\And
Reese E. Jones$^*$\thanks{\tt rjones@sandia.gov} \\
{\it Sandia National Laboratories},\\
Livermore, CA 94551
\And
Ravi G. Patel \\
{\it Sandia National Laboratories},\\
Albuquerque, NM 87185
\And
Raelynn Wonnacot \\
{\it University of Maryland},\\
College Park, MD 20742
\And
Dan S. Bolintineanu \\
{\it Sandia National Laboratories},\\
Albuquerque, NM 87185
\And
Craig M. Hamel \\
{\it Sandia National Laboratories},\\
Albuquerque, NM 87185
\And
Sharlotte L.B. Kramer \\
{\it Sandia National Laboratories},\\
Albuquerque, NM 87185
}

\ifclean
\date{}
\else
\date{\red{{\bf Do not distribute} \ Draft \today}}
\fi

\begin{document}

\maketitle

\begin{abstract}
We propose a scalable, approximate inference hypernetwork framework for a general model of history-dependent processes.
The flexible data model is based on a neural ordinary differential equation (NODE) representing the evolution of internal states together with a trainable observation model subcomponent.
The posterior distribution corresponding to the data model parameters (weights and biases) follows a stochastic differential equation with a drift term related to the score of the posterior that is learned jointly with the data model parameters.
This Langevin sampling approach offers flexibility in balancing the computational budget between the evaluation cost of the data model and the approximation of the posterior density of its parameters.
We demonstrate performance of the ensemble sampling hypernetwork on chemical reaction and material physics data and compare it to standard variational inference.

\end{abstract}

\section{Introduction}

Many models of physical and other processes require degrees of freedom to represent unobservable states that influence the observable response.
The Luenberger observer \cite{luenberger1966observers} from control theory is a widely employed example that provides an estimate of the internal state while modeling measurements of the output of a system.
Augmented neural ordinary differential equations (NODEs) \cite{chen2018neural,dupont2019augmented} is another, now ubiquitous, data representation of this type that expands the modeled dynamics to hidden dimensions for greater expressiveness via \emph{latent} variables.
Generally speaking, since the hidden variables in these types of models are inferred or postulated (as opposed to directly measured), these states should be treated as uncertain.

A wide variety of methods have been developed to perform uncertainty quantification (UQ).
Expensive, sequential Markov chain Monte Carlo algorithms \cite{geyer1992practical,gilks1995markov,brooks1998markov} are still considered the reference standards methods.
Nevertheless, more efficient optimization-based variational inference (VI) methods \cite{blei2017variational,graves2011practical} have arisen as practical alternatives.
VI methods employ a parameterized surrogate posterior and posits UQ as an optimization problem for the parameters of the surrogate.
More recently, hybrid methods that use an empirical posterior distribution composed of model realizations as well as optimization methods to drive these realizations to match data have arisen.
These include Stein VI \cite{liu2016stein} and the method we employ, Langevin sampling (LS) \cite{welling2011bayesian}.

The proposed hypernetwork framework consists of a hidden state model and an observation model that are parameterized by a sampler that is trained to data.
Unlike the widely employed Hamiltonian Monte Carlo approach \cite{girolami2011riemann,betancourt2017geometric,betancourt2017conceptual}, LS does not have a conditional acceptance criterion, which simplifies backpropagation.
The hypernetwork implementation is relatively straightforward and is fully differentiable; furthermore, LS has a firm theoretical basis which we expand upon in \aref{app:KLD}.
We illustrate the performance of the framework with process modeling applications from chemical kinetics and material physics.
After verifying the convergence properties on simpler problems, we demonstrate the efficiency and accuracy of modifying the method to treat some of the NN parameters with full UQ and others with reoptimized point estimates.

In the next section,  \sref{sec:related}, we put the proposed method in the context of related work including standard and recently developed UQ methods.
\sref{sec:method} provides the theoretical development of the algorithm and implementation details.
Then,  in \sref{sec:results}, we provide demonstrations of the method's performance in calibrating and providing UQ for selected physical models.
Finally,  \sref{sec:conclusion}, provides a summary and avenues for future work.

\section{Related work} \label{sec:related}

Uncertainty quantification (UQ) is a long-standing field of research with ongoing challenges.
Two broad classes of UQ methods are: (a) sampling-based methods, such as Markov chain Monte Carlo (MCMC) methods \cite{geyer1992practical,brooks1998markov}, which generate realizations of the posterior, and  (b) optimization-based methods, such as VI \cite{blei2017variational,graves2011practical}, which calibrate a surrogate distribution.
MCMC algorithms are challenged by high-dimensional parameter spaces, as is the case with NN models with a large number of parameters, while tractable VI methods, such as \emph{mean-field} VI \cite{blei2017variational}, suffer from limited expressiveness.

Hybrid ensemble methods have emerged that use \emph{particles} representing realizations and calibrated dynamics.
Stein VI \cite{liu2016stein} utilizes Stein's identity and a kernelized gradient flow to minimize the Kullback-Liebler (KL) divergence between the particles and the posterior underlying the data.
Like Hamiltonian Monte Carlo methods \cite{girolami2011riemann,betancourt2017geometric,betancourt2017conceptual},
Langevin sampling (LS) borrows from computational statistical mechanics \cite{frenkel2023understanding} and was proposed by Welling and Teh \cite{welling2011bayesian} as a generally applicable ensemble method bridging sampling and optimization schemes.
Later, LS was employed by \crefs{yildiz2019ode2vae,dandekar2020bayesian,xu2022infinitely} in various applications.
LS relies on a particular form of the KL divergence between two random \emph{processes} \cite{opper2019variational,li2020scalable,tzen2019neural}, as opposed to random variables/distributions.
In its simplest form, it generates samples from a Gaussian process \cite{williams1995gaussian}.
This resembles techniques such as \emph{flow matching} \cite{lipman2022flow}, which seek to minimize the distance between two random processes.
As with some diffusion models~\cite{song2020score,song2021maximum,tang2024score} and other paradigms~\cite{albergo2023stochastic,chandramoorthy2024score,chandramoorthy2024learning}, it is based on learning a \emph{score} function.
Furthermore, methods have been designed that have  strongly interacting Langevin diffusions \cite{garbuno2020interacting}, unlike the approach adopted in this work.

We formulate our LS method as a hypernetwork, where the learned score-based sampler provides parameters to the data model, which produces outputs that are compared to data.
Hypernetworks have found  wide-spread use in a variety of tasks \cite{ha2016hypernetworks,chauhan2024brief, krueger2017bayesian, johnson2013hypernetworks, beck2023hypernetworks, wang2010evolving, zhang2018graph, lorraine2018stochastic, zhao2020meta}.
Our utilization of a divided parameter space, where one partition is optimized to improve fit while the other pursues the UQ objective,
resembles \emph{optimal brain surgery} \cite{hassibi1992second,hassibi1993optimal}, a deterministic method for model-parameter pruning.

\section{Methodology} \label{sec:method}

The goal of this work is to provide comprehensive uncertainty quantification (UQ) of a class of generally applicable NN models.
We combine stochastic sampling of a data matching model in a form of a hypernetwork where a trainable sampler provides weights to the data representation model whose predictions are compared to data.

\begin{figure}
\centering
\begin{subfigure}[b]{0.65\textwidth}
\includegraphics[width=0.95\textwidth]{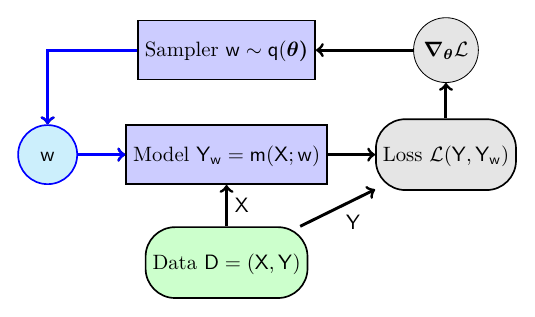}
\caption{Hypernetwork}
\end{subfigure}
\begin{subfigure}[c]{0.45\textwidth}
\includegraphics[width=0.95\textwidth]{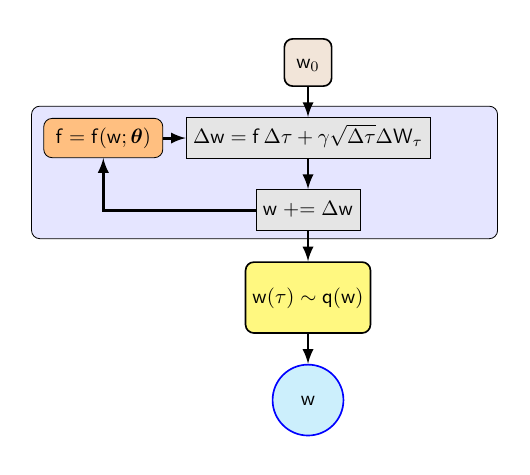}
\caption{Sampler subcomponent}
\end{subfigure}
\begin{subfigure}[c]{0.45\textwidth}
\includegraphics[width=0.95\textwidth]{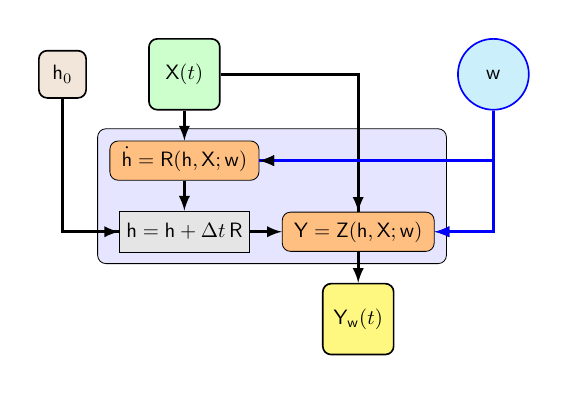}
\caption{Data model subcomponent}
\end{subfigure}
\caption{Schematics: (a) hypernetwork consisting of subcomponents:
(b) sampler $\weights \sim \surrogateposterior(\weights | \model, \data ; \parameters)$,
and (c) data model $\outputs = \model(\inputs; \weights)$
}
\label{fig:hypernet}
\end{figure}

\paragraph{Hypernetwork framework}
The framework (refer to the  schematic in \fref{fig:hypernet}a) consists of data $\data$, a data model $\model$, a loss $\loss$, and a parameter sampler $\surrogateposterior$.
Data consists of input/output pairs $\data = \{ \inputs, \outputs \}$.
The data model $\model$ takes weights $\weights$ from the sampler and data inputs $\inputs$ to produce outputs $\outputs$
\begin{equation} \label{eq:data_model}
\outputs_\weights = \model(\inputs; \weights).
\end{equation}
The sampler is a generative model for the weights $\weights$ of model $\model$
\begin{equation}
\weights \sim \surrogateposterior(\weights \ | \ \model, \data; \parameters),
\end{equation}
which is parameterized by $\parameters$,
and relies on a stochastic differential equation (SDE):
\begin{equation} \label{eq:sampler_sde}
\dd \weights = \drift(\weights; \parameters) \dd \tau + \diffusion \sqrt{2} \dd \Bs_\tau
\end{equation}
to generate samples of the surrogate posterior $\mathsf{q}$.
In \eref{eq:sampler_sde}, $\drift$ is the \emph{drift}, parameterized by $\parameters$, and is tuned to data via a loss metric, and the Brownian motion $\dd \Bs_\tau$ drives diffusion.
Note, $\sqrt{2}$ is a normalizing constant, see App.~\ref{app:OU} for more explanation.
Together the weight sampler, \eref{eq:sampler_sde} and the data model, \eref{eq:data_model}, form a \emph{hypernetwork} \cite{ha2016hypernetworks,chauhan2024brief}, whereby one network generates the weights for the another.

\paragraph{Data}
The data $\data$ we focus on consists of observables that evolve in time and can depend on the previous history of the inputs.
This type of data is typical of the mechanical response of materials, reaction kinetics and other physical systems with hidden/internal states.

\paragraph{Data model}
The data model $\model$, refer to \fref{fig:hypernet}c, consists of (a) an ordinary differential equation (ODE) to evolve the hidden state
\begin{equation} \label{eq:hidden_model}
\dot{\hiddenstate}
= \rhs(\hiddenstate,\inpt(t))
= \NN_\rhs(\hiddenstate,\inpt(t); \wH)
\end{equation}
and (b) an observation model
\begin{equation} \label{eq:obs_model}
\outputs_\weights
= \observation(\hiddenstate,\inpt(t))
= \NN_\observation(\hiddenstate,\inpt(t); \wO)
\end{equation}
where $\hiddenstate$ is the hidden state with presumed initial condition $\hiddenstate_0 = \mathbf{0}$ (unless there is an explicit initial state), $\inpt \in \inputs$ is a given time-dependent input, and $\outputs$ are the corresponding observable outputs.

\paragraph{Sampler}
The hypernetwork is a probabilistic modeling framework to estimate the model parameters, $\weights=\{\wH,\wO\}$.
In a Bayesian context, the posterior distribution $\prob(\weights\vert\data)$ is given by:
\begin{equation}
\prob(\weights\vert\data)\propto \prob(\data\vert \weights)\prob(\weights)
\label{eq:bayes}
\end{equation}
where $\prob(\data\vert \weights)$ is the likelihood of observing the data $\data$ given the model $\model$ parameterized by $\weights$, and $\prob(\weights)$ is the prior distributions for the model parameters.

The posterior distribution $\prob(\weights\vert\data)$ is not, in general, available analytically.
In this work, we use LS via the SDE \eqref{eq:sampler_sde} to provide samples of the posterior.
\fref{fig:hypernet}b provides a schematic of the sampler.

We connect the score-based drift $\drift = \grad_\weights \PE$ in \eqref{eq:sampler_sde} with a
\emph{potential energy} $\PE(\weights)$ associated with the model (negative log) posterior/\emph{score}
\begin{equation}
\PE(\weights) = -\log \prob(\weights\vert\data)=- \log \prob(\data\vert \weights)-\log \prob(\weights) + \text{constant}
\label{eq:potenrg}
\end{equation}
so that
\begin{equation}
\grad_\weights \PE(\weights) =
- \grad_\weights \left[ \log \prob(\data\vert \weights)+\log \prob(\weights) \right].
\end{equation}
With this connection, we have an SDE that samples the posterior of the weights of the data model:
\begin{equation} \label{eq:langevin_dyn}
\dd\weights = -\grad_\weights \PE(\weights) \dd \tau + \sqrt{2} \, \dd \Bs_\tau
\end{equation}
where $\Bs_\tau \in \RE^{|\weights|}$ is a standard multidimensional Brownian motion/Weiner process, with $|\weights|$ denoting the cardinality of $\weights$, and $\tau$ pseudo-time.

\paragraph{Weight model}
Following \crefs{welling2011bayesian,xu2022infinitely,li2020scalable,tzen2019neural}, we approximate  the prior $\prob(\weights)$  and the posterior $\prob(\weights \vert \data)$  distributions with generative SDEs:
\begin{eqnarray}
\dd \weights &=& \drift_{\priortag}(\weights) \dd \tau + \diffusion_{\priortag} \sqrt{2} \dd \Bs_\tau\hspace{0.2in}\textrm{(prior)} \label{eq:sdeprior}\\
\dd \weights &=& \drift_{\posteriortag}(\weights) \dd \tau + \diffusion_{\posteriortag} \sqrt{2} \dd \Bs_\tau\hspace{0.2in}\textrm{(posterior)}\label{eq:sdepost}
\end{eqnarray}
where $\drift_{\posteriortag} = \grad_\weights \prob(\weights \vert \data)$ and $\drift_{\priortag} = \grad_\weights \prob(\weights)$ are the drift components for the posterior and prior, respectively,  and $\diffusion_{\posteriortag}$ and $\diffusion_{\priortag}$ are the corresponding diffusion components magnitudes.

As in \cref{xu2022infinitely}, we employ a Ornstein-Uhlenbeck (OU) process for the prior with $\drift_{\priortag}(\weights)=-(\weights-\bar{\weights})$ and set $\diffusion_{\priortag}=\diffusion_{\posteriortag}=1$ and $\dd\Bs_\tau\sim\mathcal{N}(\mathbf{0},\dd\tau\Is)$.
The steady state OU distribution is then a multivariate normal with diagonal unit covariance, $\weights \sim \Nc(\bar{\weights},\Is)$, and mean $\bar{\weights}$ which is set to zeros, the maximum likelihood estimate (MLE) or other point estimate.
The score of the posterior $\grad_\weights \prob(\weights \vert \data) $ is approximated by a neural network, $\NN_\drift(\weights;\parameters)$, where $\parameters$ are the parameters defining the posterior model
\begin{equation}
\drift_\posteriortag = \NN_\drift(\weights ; \parameters) \ .
\end{equation}

\paragraph{Loss}
We employ the evidence lower bound (ELBO)~\cite{blei2017variational} to construct the loss function that will drive the search for an approximate posterior distribution for the model parameters $\weights$
\begin{equation} \label{eq:elbo}
\loss_\text{ELBO} =
\Exp_{\surrogateposterior(\weights;\theta)}\left[\log\prob(\data\vert \weights)\right] -\KLD(\surrogateposterior(\weights;\theta)\vert\vert \prob(\weights)) \ ,
\end{equation}
where $\KLD(\surrogateposterior(\weights;\theta)\vert\vert \prob(\weights))$ is Kullback-Liebler (KL) divergence between the prior and the approximate posterior of the weights $\weights$ of the physical process model.
The first term promotes accuracy by informing the surrogate posterior $\surrogateposterior$ by the likelihood $\prob(\data | \weights)$, and the second term acts as a regularizing penalty by incurring a cost for the surrogate posterior being far from the prior $\prob(\weights)$.
If both SDEs for the prior and posterior distributions, in Eqs.~\eqref{eq:sdeprior} and \eqref{eq:sdepost}, share the same diffusion, $\diffusion = \diffusion_{\priortag}=\diffusion_{\posteriortag}$, the KL divergence between the prior and the posterior is finite~\cite{tzen2019neural}, and the KL divergence between the two processes can be written as:
\begin{equation}
\KLD(q(\weights;\theta)\vert\vert \prob(\weights))
=\Exp_{q(\weights;\theta)}\left[\frac{1}{2}\int_0^{\tau_f} \left\| \frac{ \drift_{\posteriortag}(\weights_\tau)-\drift_{\priortag}(\weights_\tau)}{\diffusion}\right\|^2_2\, \dd \tau\right].
\label{eq:kldivint}
\end{equation}
This expression is an expectation over the Langevin samples generated by the numerical solution of the SDE model.
We select the $\weights$ samples corresponding to the pseudo-time $\tau_f$ as an approximation of sampling from $\surrogateposterior(\weights;\parameters)$ for the purpose of estimating the marginal likelihood
\begin{equation}
\Exp_{q(\weights;\theta)}\left[\log\prob(\data\vert \weights)\right]\approx\frac{1}{N_s}\sum_{i=1}^{N_s} \log\prob(\data\vert \weights_i)
\end{equation}
where $\weights_i=\weights_i(\tau_f;\parameters)$.
Refer to \aref{app:KLD} for a brief derivation of this form of the KL divergence and how it bounds the KL divergence in \eref{eq:elbo} for the distribution of weights at $\tau_f$.

In this work, we consider time-dependent processes, with model outputs
\begin{equation}
\outputs_{\weights}=\{\outputs_{\weights}(X_j,t_k; \weights_i)\vert i=1,\ldots,N_s, j=1,\ldots N_D, k=1,\ldots,N_t\}
\end{equation}
collected at a sequence of (physical) time steps, $t_1,\ldots,t_{N_t}$ and realizations of the weight samples $\weights_i$, $i=1,\ldots,N_s$.
Given the set of model evaluations, we construct the likelihood $\prob(\data\vert \weights)$ as follows.
We consider data realizations that were generated independently, thus
\begin{equation}\label{eq:likelihood}
\prob(\data\vert \weights_i)=\prod_{j=1}^{N_D}\prob(\data_j\vert \weights_i).
\end{equation}
For both numerical and computational experiments, this is a reasonable approximation as data is typically collected independently for different experimental designs.
Further, we approximate independent observations across each time series, leading to
\begin{equation}
\prob(\data\vert \weights_i)=\prod_{j=1}^{N_D}\prod_{k=1}^{N_t}\prob(\outputs_{j,k}\vert \weights_i)
\end{equation}
where $\data=\{\outputs_{j,k}\vert j=1,\ldots N_D, k=1,\ldots,N_t\}$.
This approximation can be relaxed if correlation information is readily available or if sufficient data is available to estimate it numerically, e.g. via entropy-based methods.

Furthermore, we employ a Gaussian mixture model for $\prob(\outputs_{j,k}\vert \weights_i)$
\begin{equation}
\prob(\outputs_{j,k}\vert \weights_i) = \sum_{l=1}^{N_\text{modes}} \alpha_l \, g(\outputs_{j,k}; \outputs_{w_i;j,k,l}, \sigma_{k,l})
\end{equation}
where
\begin{equation}
g(\outputs_{j,k}; \outputs_{\weights_i;j,k,l}, \sigma_{k,l})\propto
\frac{1}{2\sigma_{k,l}^2}
\exp\left(-\frac{(\outputs_{j,k}-\outputs_{\weights_i;j,k,l})^2}{2\sigma_{k,l}^2}\right).
\end{equation}
The results presented in \sref{sec:results} employ either one or two modes, $N_\text{modes}=1,2$.
For the results corresponding to $N_\text{modes}=2$, the modes were sufficiently distinct to estimate the mode weights $\alpha_l$ by counting the number of data trajectories that fall into either mode.
In a more generic setting, with multiple and/or less distinct modes, one could use techniques such as k-means clustering to estimate this information.

The standard deviations, $\sigma_{k,l}$, can either be added to the collection of parameters inferred along with the model parameters or can be estimated by other means.
Here, we estimate this value as
\begin{equation}
\sigma_{\cdot}^2=\sigma_{\cdot,\data}^2+\sigma_{\cdot,\text{MLE}}^2
\end{equation}
where $\sigma_{\cdot,\data}$ represents the standard deviation computed using the data only, while $\sigma_{\cdot,\text{MLE}}$ represents the discrepancy between the mean data and the model fitted via MLE.
This expression assumes that the data generating process and the process of fitting a model contribute to uncertainty in the outputs.
In practice, however, we found that $\sigma_{\cdot,\text{MLE}}$ was much smaller than the data counterpart for all models presented in this paper.

\paragraph{Algorithm}
We collect the relevant equations, Eqs.~\eqref{eq:data_model}, \eqref{eq:sampler_sde}, and \eqref{eq:hidden_model}, into the framework:
\begin{eqnarray}
\dd\weights &=& \NN_\weights(\weights; \parameters) \dd \tau + \diffusion\,\sqrt{2}\dd \Bs_\tau  \nonumber \\
\dd \hiddenstate &=& \NN_\rhs(\inputs,\hiddenstate;\wH) \dd t \label{eq:hypernet_framework} \\
\outputs_\weights &=& \NN_\observation(\inputs,\hiddenstate;\wO) \nonumber.
\end{eqnarray}
The algorithm starts by first fitting $\outputs_\weights$ to data using the likelihood in \eref{eq:likelihood} for the discrepancy between the output of the data model in \eref{eq:hypernet_framework}$_\text{b,c}$ and the available data.
This results in $\weights_\text{MLE}=\{\wH,\wO\}$ that will be used subsequently in the initial conditions $\weights_0$ for the SDE model of the weight posterior distribution.
The framework then proceeds to optimize parameters $\parameters$ in the following manner:
\begin{itemize}
\item Generate initial conditions for several SDE trajectories by perturbing the MLE solution for $\weights$; generate several weight trajectories in pseudo-time $\tau$ using an Euler-Maruyama integrator
\begin{equation}
\weights_{i+1} = \weights_{i} +  \drift(\weights_i,t_i) \Delta\! \tau + \diffusion\sqrt{2}\Delta\Bs_\tau.
\end{equation}
Here the diffusion term contribution is given by $\Delta\! \Bs_\tau\sim\mathcal{N}(\mathbf{0},\Delta\! \tau\Ib)$.
For the results presented in \sref{sec:results}, without loss of generality, we set $\diffusion = 1$.
This is motivated by the fact that, when the diffusion term correspond to $\sqrt{2}\Delta\! \Bs_\tau$, samples are distributed according to the posterior distribution corresponding to the score function $\drift(\weights_i,t_i)$; see \aref{app:OU}.
For some demonstrations, however, we did find that initially annealing the strength of the prior diffusion with  $\gamma_0 > 1$ did have advantages in obtaining a diverse posterior.
\item Collect parameters $\weights$ at the end of all trajectories, and feed them into \eref{eq:hypernet_framework}$_\text{b,c}$ for $\hiddenstate$ and $\outputs$, respectively.
For the ODE corresponding to $\hiddenstate$, we employ the explicit, predictor/corrector Heun integrator
\begin{eqnarray}
\hiddenstate_{n+1}^{(p)} &=& \hiddenstate_{n} + \Delta t \, \NN_\rhs(\inpt_{n+1},\hiddenstate_{n}) \\
\hiddenstate_{n+1} &=& \hiddenstate_{n} +
\frac{1}{2} \left(\NN_\rhs(\inpt_{n+1},\hiddenstate_{n})+\NN_\rhs(\inpt_{n+1},\hiddenstate_{n+1}^{(p)})\right).
\end{eqnarray}
The size of the hidden state $\hiddenstate$ is a hyperparameter and the initial condition is set to $\hiddenstate_0 = \mathbf{0}$, unless otherwise available in a given system.
The observation model $\NN_\observation$ is evaluated at discrete (physical) timesteps $t=\{t_0,t_1\ldots t_n\}$, corresponding to time steps where observations are available.
\item Evaluate the ELBO loss function~\eqref{eq:elbo} and its gradients, and update $\parameters$ using a typical optimization scheme, e.g. ADAM~\cite{kingma2014adam}.
\end{itemize}
We also employ a hybrid framework inspired by the \emph{Bayesian last layer} (BLL) work~\cite{lazaro2010marginalized,watson2021latent}.
In this version of the  BLL framework, we designate subsets of parameters that will be considered stochastic, typically the last layer or last few layers in $\NN_\rhs$ and $\NN_\observation$.
This choice is motivated by the observation that the layers closer to the input adapt the space to match the input-output dynamics, while the last set of layers construct a representation for the output in this transformed space.
Formally, we designate
\begin{equation}
\weights=
\underbrace{\{\wH^{(d)},\wO^{(d)}\}}_{\weights^{(d)}}
\cup
\underbrace{\{\wH^{(s)},\wO^{(s)}\}}_{\weights^{(s)}}
\end{equation}
where the superscript $d$ corresponds to the (deterministic) MLE optimization set while $s$ corresponds to the stochastic set.
For this set, the model training will jointly optimize parameter subsets, $\weights^{(d)}$, and $\parameters$ corresponding to the score function that is used by the Langevin sampler to generate $\nrepl$ replicas $\weights_j^{(s)}=\{\wHj^{(s)},\wOj^{(s)}\}$ with $j=1,\ldots,\nrepl$.

While a single SDE trajectory could, in principle, be sufficient to generate statistical samples of model parameters $\weights$, here we employ $\nrepl$ replica trajectories for two reasons:
(1) simulating trajectories is an inherently parallel process and thus one only needs to advance trajectories until they reach a steady state and collect the last state as a sample, and (2) the numerical estimate of the KL divergence term in \eref{eq:kldivint} relies on the availability of several SDE trajectories.

\Aref{alg:training} summarizes the overall procedure.

\begin{algorithm}[htb]
\begin{algorithmic}[1]
\State Fit MLE model $\model =$ ($\NN_\rhs$, $\NN_\observation$) to data $\data = (\inputs, \outputs)$ to obtain $\weights^\text{(MLE)}$
\State Initialize the Langevin sampler.
\begin{itemize}
\item Instantiate $\nrepl$ initial conditions for the Langevin sampler: $\weights_{0,j}\sim \mathcal{N}(\weights^\text{(MLE)},\epsilon\Is)$, $j=1,\ldots,\nrepl$ and $\epsilon$ a small value (we used $10^{-5}$).
\item Instantiate parameters $\parameters$ for drift term of SDE $\NN_\weights$
\end{itemize}
\For {$epoch = 1,\ldots,N_{\text{epochs}}$}
\State Generate $\nrepl$ trajectories $\weights_j^{(s)}(\tau)$, $\tau=0\ldots\tau_f$
\State Assemble $\weights_j=\{\weights^{(d)},\weights_j^{(s)}\}$ and compute $\outputs_j=\model(\inpt,\weights_j)$, $j=1,\ldots,\nrepl$.
\State Compute the loss function Eq.~\ref{eq:elbo} via numerical expectation estimates given $\nrepl$ set of parameters.
\State Compute gradients of loss function and update $\weights^{(d)}$ and $\parameters$.
\EndFor
\end{algorithmic}
\caption{Training algorithm}
\label{alg:training}
\end{algorithm}

In the next section we will compare inference results obtained with the algorithm proposed above with results obtained via variational inference~\cite{Jordan:1999}.
Specifically, we employ the \emph{black-box variational inference} (BBVI) method for this purpose~\cite{Ranganath:2014}.

\section{Demonstrations} \label{sec:results}

To demonstrate the versatility of the proposed method, we use synthetic data with various amounts of stochastic noise and/or internal variance.
First, in Sec.~\ref{sec:res_chem}, we illustrate the performance of the UQ scheme with data generating models from reaction chemistry.
Systems of SDEs are widely used in modeling chemical (and physical) processes \cite{lemons2002introduction,coffey2012langevin,adelman1980generalized}, and we use exemplars to explore the performance of the algorithm to combinations of external noise and internal variations that result in epistemic and aleatory uncertainty.
Then, in Sec.~\ref{sec:matphys}, we turn to applications from material physics, specifically a viscoelastic composite with heterogenities that are treated as uncertainty around the mean response.
For this application we have no closed form truth model.
We compare the proposed LS method to BBVI for the chemical reaction demonstrations, and with the material physics demonstrations we contrast various Bayesian last layer adaptations of the LS method.

We use the Wasserstein 1-distance ($\Wc_1$) to measure the similarity of two probability densities $\prob_a$ and $\prob_b$:
\begin{equation} \label{eq:W1}
\Wc_1(\prob_a,\prob_b)
= \int_{-\infty}^\infty \left| \CDF_a(y) - \CDF_b(y) \right| \, \mathrm{d}y,
\end{equation}
using cumulative distribution functions (CDFs) which we construct empirically from samples.

\subsection{Chemical kinetics} \label{sec:res_chem}
We use two physical SDEs to demonstrate the convergence properties of the proposed method.
The first is classical Langevin dynamics.
Langevin dynamics are used in a variety of chemical models from thermostatted molecular  dynamics \cite{frenkel2023understanding,hunenberger2005thermostat,toton2010temperature} to Mori-Zwanzig-based multiscale models \cite{adelman1974generalized,adelman1976generalized,wagner2003coupling}.
Herein we employ a simple Langevin system that has an exact solution so we can investigate a case with no model discrepancy.
Then we turn to a stochastic model of the \Schl \ reaction that exhibits bifurcations and therefore multimodal response density functions.

\paragraph{Langevin dynamics}

The goal of this demonstration is to: (a) show convergence of the proposed method to exact results with a NN that is consistent with the data-generating model, and (b) to show behavior of the method in the presence of internal (embedded) data-generating model variability \cite{sargsyan2015statistical}.

The data-generating model is a 1D Langevin equation
\begin{equation} \label{eq:1D_OU_sde}
\dd y = -\gamma (y - \ybr) \dd t + \sigma \dd B_t
\end{equation}
where $\gamma$ and $\sigma$ are parameters controlling the decay to the mean $\ybr$ and the magnitude of the random force.
To generate data, we choose $\gamma \sim \Nc(8,0.8)$, $\ybr \sim \Nc(1,0.1)$, initial conditions $y_0=y(t=0)\sim \Nc(2,0.02)$, and $\sigma=0$.
For each tuple $(\gamma,\ybr,y_0)$, the model admits a analytical solution,
\begin{equation*}
y(t) = (y_0-\ybr)\exp(-\gamma t)+\ybr
\end{equation*}

We model the process data with a reduction of the proposed hypernetwork framework:
\begin{eqnarray} \label{eq:langevin_snode}
\dd \weights &=& \NN_\weights(\weights ; \parameters) \, \dd \tau + \sqrt{2} \dd \Bs_\tau \nonumber \\
\dd \hiddenstate &=& \NN_\rhs(\hiddenstate; \weights) \, \dd t  \\
\outpt &=&  \hiddenstate \nonumber
\end{eqnarray}
where the observation model is the identity $\outpt \equiv \hiddenstate$ and $\NN_\rhs$ just needs to learn a linear function of $\hiddenstate$.
To create a data model consistent with \eref{eq:1D_OU_sde}, we used a two-parameter NN for $\NN_\rhs(\hiddenstate; \weights)$, i.e. a linear model:
\begin{equation}
\NN_\rhs(y;\{W,b\})  \equiv W y + b
\end{equation}
Given the model in \eref{eq:1D_OU_sde}, with $W=-\gamma$ and $b=\gamma\ybr$, the theoretical summary statistics of the two NN parameters are:
\begin{eqnarray} \label{eq:langevin_weight_correlation}
\Exp[W] &=& -\Exp[\gamma] = -8.0  \\
\Exp[b] &=& \Exp[\gamma \ybr] = \Exp[\gamma] \Exp[\ybr] = 8.0 \nonumber\\
\Var[W] &=&  \Var[\gamma] = 0.64 \nonumber\\
\Var[b] &=&  \Var[\gamma]\Var[\ybr]+ \Exp[\gamma]^2\Var[\ybr] + \Var[\gamma]\Exp[\ybr]^2 = 1.29 \nonumber\\
\Cov[W,b]
&=& -(\Exp[\gamma^2\ybr] - \Exp[\gamma]\Exp[\gamma\ybr])
= -\Var[\gamma] \Exp[\ybr]
= -0.64 \nonumber\\
\rho[W,b] &=& \Cov[W,b]/\sqrt{\Var[W]\Var[b]}=-0.71\nonumber
\end{eqnarray}
Note, \aref{app:LScore} provides a derivation of the score function for this demonstration.

\begin{figure}[htb]
\centering
\begin{subfigure}[c]{0.32\textwidth}
\includegraphics[width=0.99\linewidth]{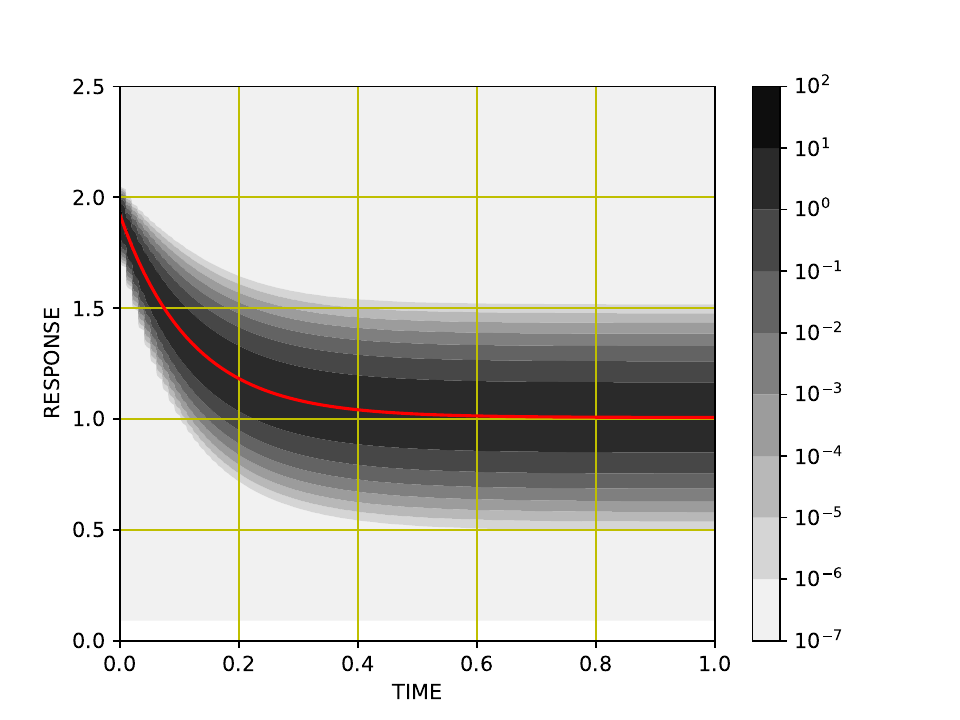}
\caption{data density}
\end{subfigure}
\begin{subfigure}[c]{0.32\textwidth}
\includegraphics[width=0.99\linewidth]{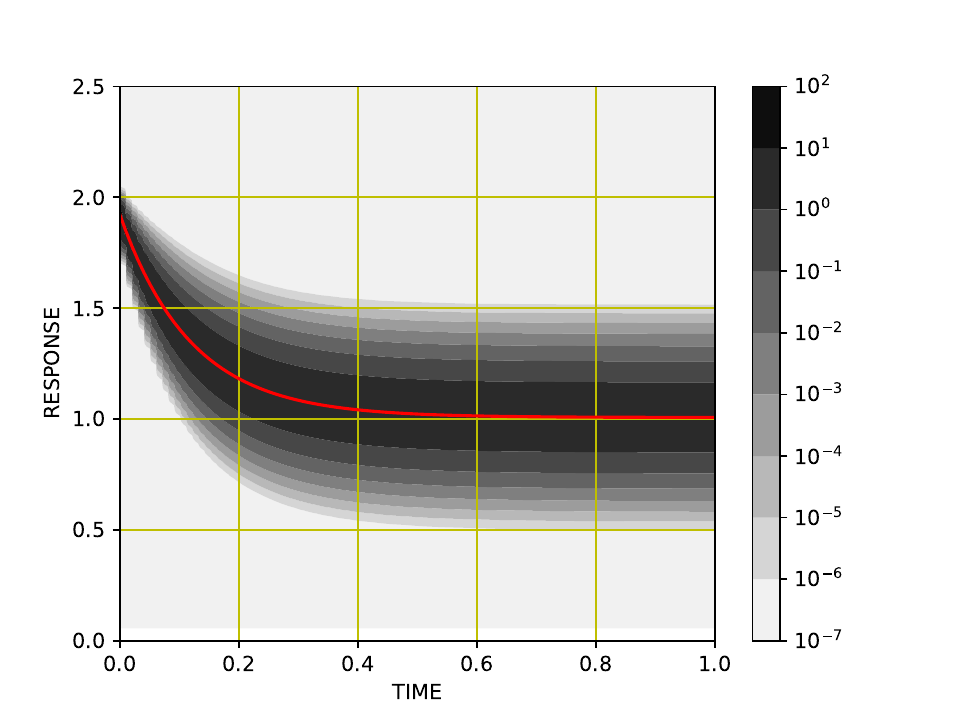}
\caption{predicted density via LS (\Aref{alg:training})}
\end{subfigure}
\begin{subfigure}[c]{0.32\textwidth}
\includegraphics[width=0.99\linewidth]{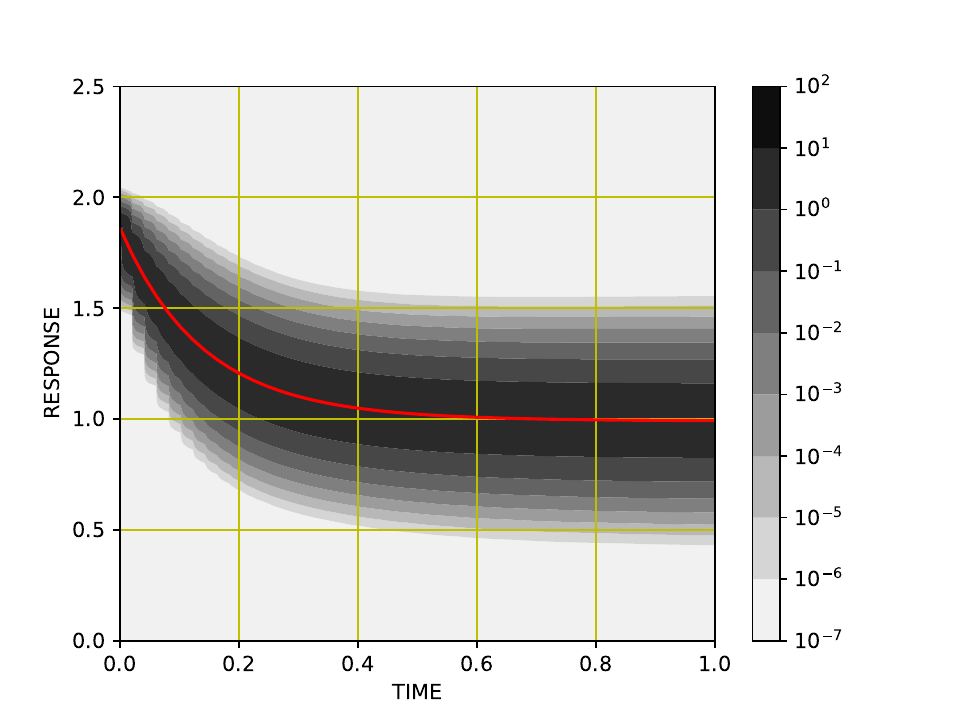}
\caption{predicted density via BBVI}
\end{subfigure}
\caption{Langevin dynamics: (a) data density, (b) predicted density via Langevin sampling (Algorithm~\ref{alg:training}), and (c) predicted density via BBVI. The mean trend is shown with a red line.
}
\label{fig:langevin_trajectories}
\end{figure}

For this demonstration, we used a dataset with 1024 trajectories and employed the last weight samples of $10^4$ pseudotime steps.
For the Langevin sampling approach, the drift NN for the sampler $\NN_\weights$ had 2 internal layers with 2 nodes each and $\tanh$ activations.
For the BBVI approach we adopted a (full covariance) bivariate Gaussian distribution model that consists of means and standard deviations for both $W$ and $b$, as well as the correlation coefficient between the two, $\rho[W,b]$, for a total of 5 parameters.
\fref{fig:langevin_trajectories} shows kernel density estimates (KDE) for the data and the two variational inference exercises.
The KDE results employ 1024 trajectories each, and the estimates are computed discretely at 100 intermediate times.
There are no visible difference between the data, shown in the left frame, and the Langevin sampling results, in the middle frame.
Although the uncertainties seemed well-captured by the BBVI approach (shown in the right frame) in the asymptotic regime towards the end of the time axis, at early time the BBVI results are clearly different compared to the original data.

\begin{figure}[htb]
\centering
\includegraphics[width=0.75\linewidth]{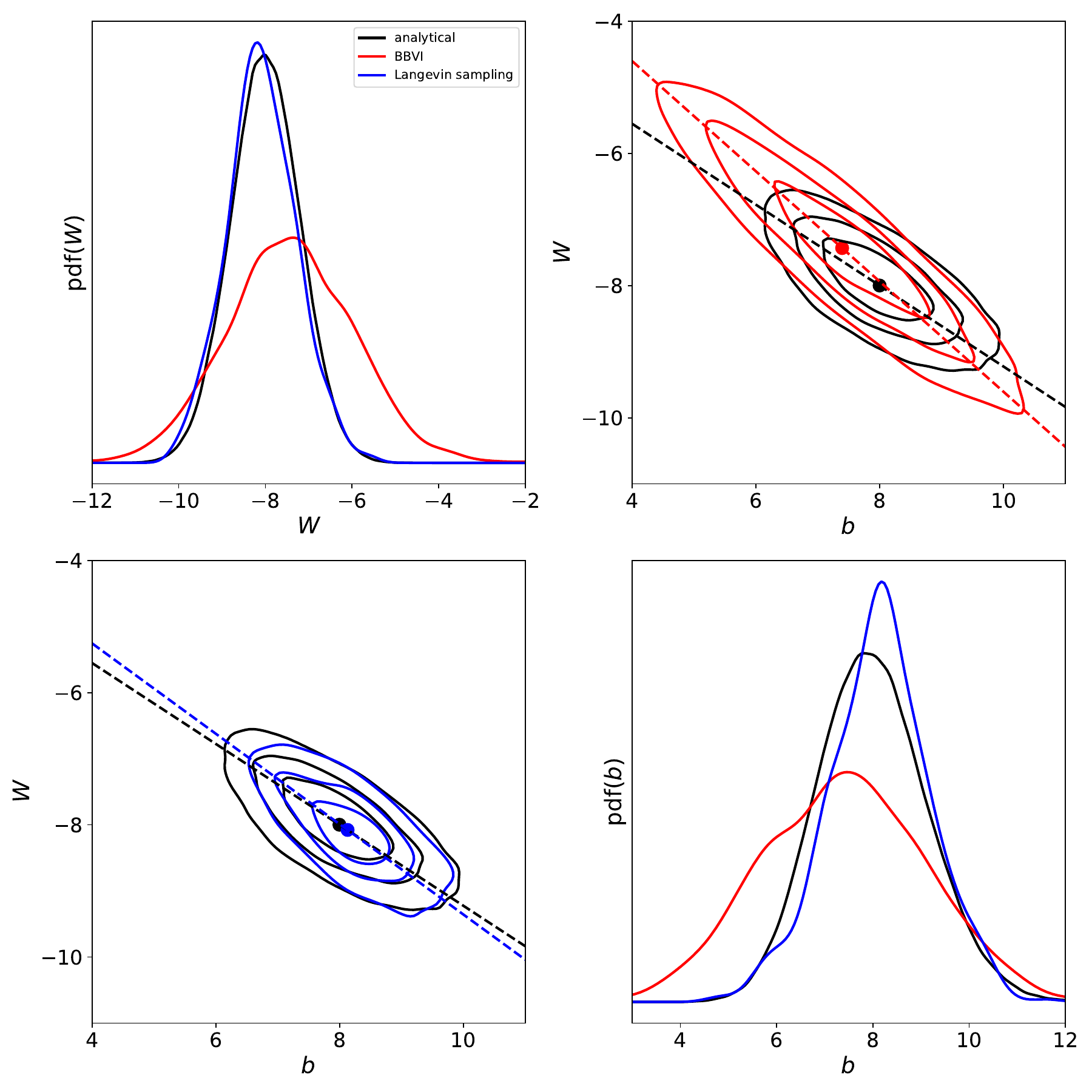}
\caption{Langevin model: kernel density estimates of 1D marginal distributions and 2D joint densities for model parameters $(W,b)$.
The dashed red and blue lines shown in the 2D density plots represent the first principal vector constructed from available $(W,b)$ for BBVI (upper right panel) and Langevin sampling (lower left panel), respectively,  and the dashed black line represents the theoretical solution (in both off diagonal panels).
}
\label{fig:langevin_correlation}
\end{figure}

\fref{fig:langevin_correlation} depicts the 1D marginal and 2D joint densities of $\weights=\{W,b\}$.
The 2D join densities also feature the corresponding first principal vectors passing through the mean values shown with large circles with colors matching the scheme employed these plots.
The Langevin sampling results (blue) are close to the analytical model statistics shown in black, while the BBVI (red) shows significant discrepancies.
The Langevin sampling standard deviation for $W$ matches the theoretical value while the corresponding value for $b$ is within 7\% of the analytical result.
The BBVI results for the standard deviations, $(1.46,1.73)$ are about 70\% larger.
Despite the larger parameter uncertainties displayed by the BBVI results it seems the agreement in predicted uncertainties are achieved by adjusting the correlation structure, as evidenced by the results shown in the top right frame.
BBVI displays a larger correlation coefficient between the parameters, approxmately $0.92$, compared to $0.71$ in the original model.
The comparison shown in the lower left plot show that the proposed LS approach captures well the correlation between model parameters, with just a small discrepancy between the slopes of the corresponding principal component vectors.

\begin{figure}[htb]
\centering
\includegraphics[width=0.45\linewidth]{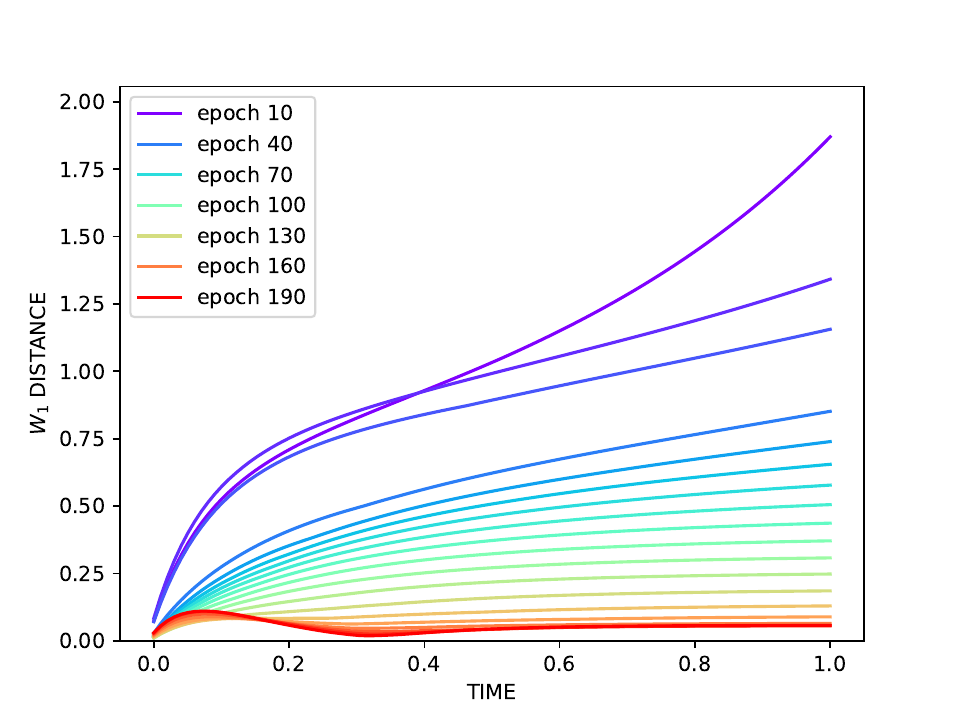}
\caption{Langevin dynamics:
Wasserstein $\mathcal{W}_1$ distance as function of time and epochs  for sequence of samples. }
\label{fig:langevin_convergence}
\end{figure}

Lastly,
\fref{fig:langevin_convergence} illustrates the convergence of the predicted response distribution via Langevin sampling compared to the data distribution using the Wasserstein 1-distance ($\Wc_1$), in \eref{eq:W1}.
At early epochs the largest $\Wc_1$ occur at late times.
This is to be expected, since small model inaccuracies can lead to trajectories that depart from the asymptotic behavior of the actual model.
The errors appear to reduce proportionally until later epochs, where some relatively larger discrepancies are observed at early times, likely to accommodate smaller discrepancies over the remaining of the time span.

\paragraph{Schl\"{o}gl model}
The goal of this demonstration is to show the performance of the proposed algorithm in the presence of high-frequency noise in the physical response and a multi-modal posterior.
For this, we employ Schl\"{o}gl’s model, which is a canonical chemical reaction network model that exhibits bistability~\cite{Schlogl:1972,vellela2009stochastic,sargsyan2010spectral}.

\begin{figure}[htb]
\centering
\begin{subfigure}[b]{0.3\textwidth}
\includegraphics[width=0.99\linewidth]{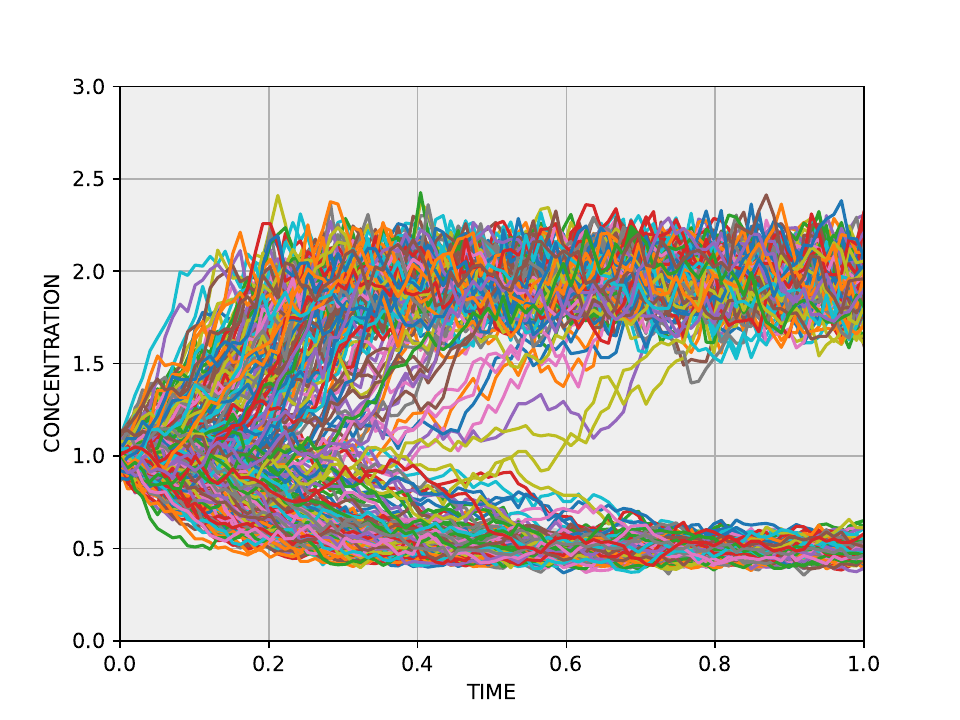}
\caption{data trajectories }
\end{subfigure}
\begin{subfigure}[b]{0.3\textwidth}
\includegraphics[width=0.99\linewidth]{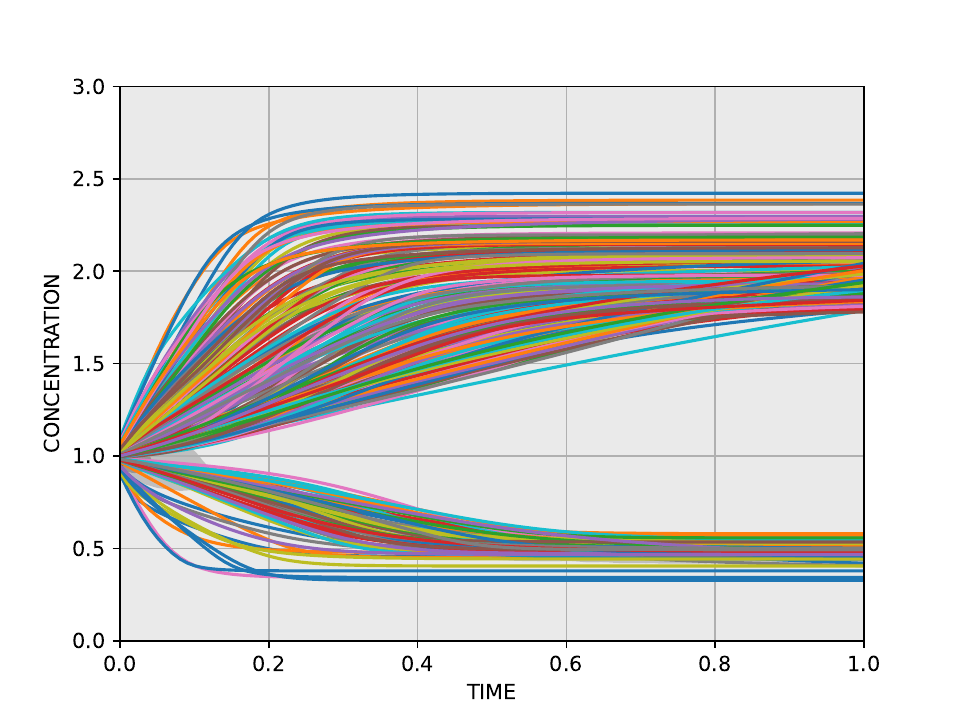}
\caption{LS trajectories}
\end{subfigure}
\begin{subfigure}[b]{0.3\textwidth}
\includegraphics[width=0.99\linewidth]{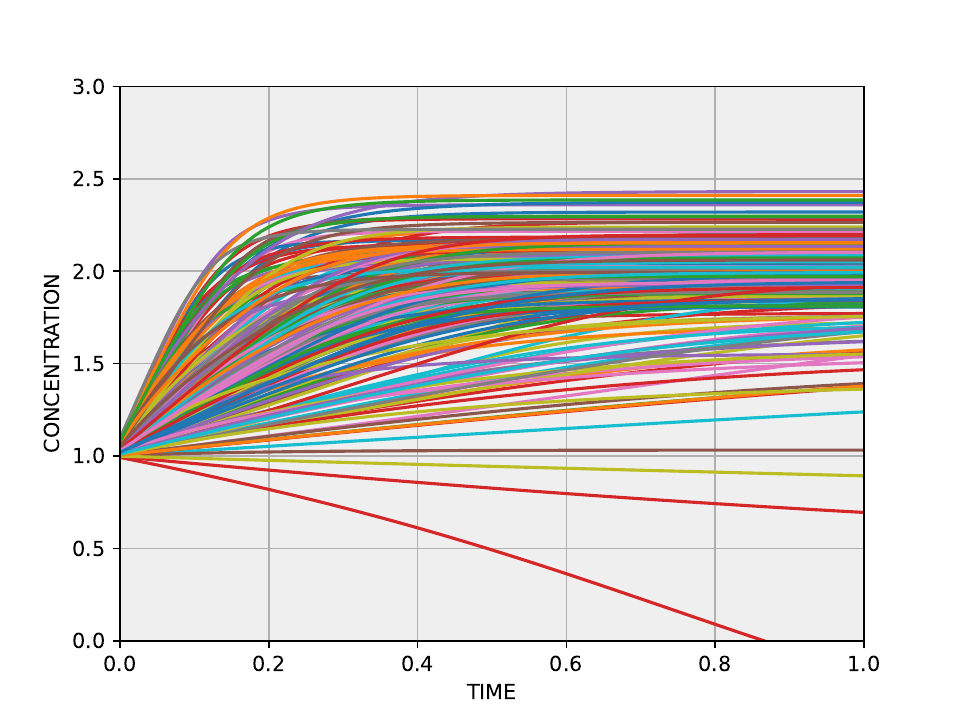}
\caption{BBVI trajectories }
\end{subfigure}
\begin{subfigure}[b]{0.3\textwidth}
\includegraphics[width=0.99\linewidth]{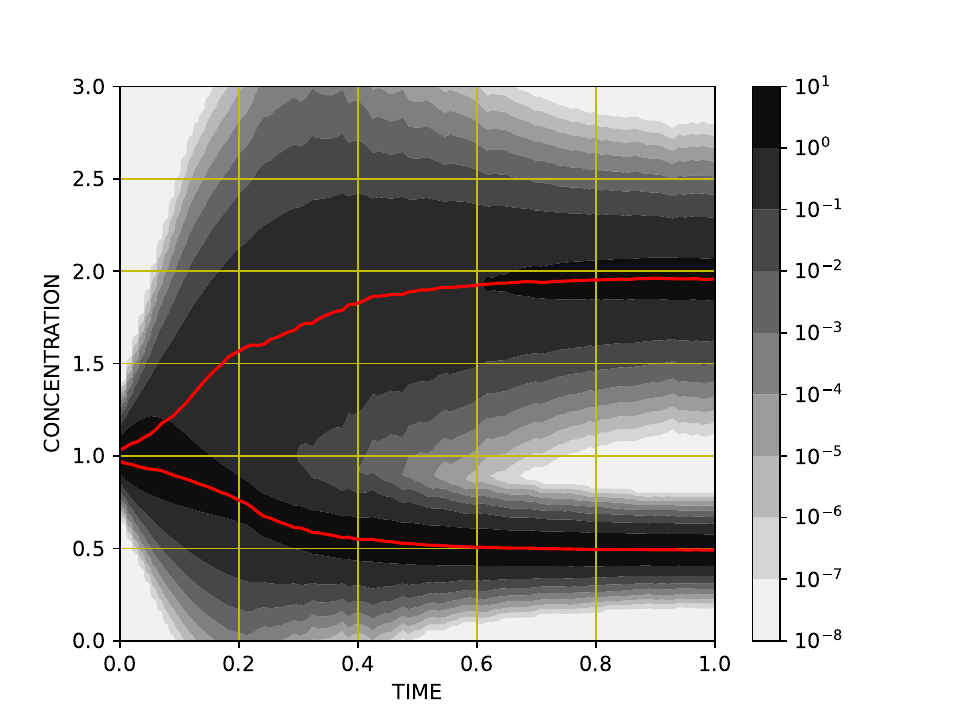}
\caption{data density}
\end{subfigure}
\begin{subfigure}[b]{0.3\textwidth}
\includegraphics[width=0.99\linewidth]{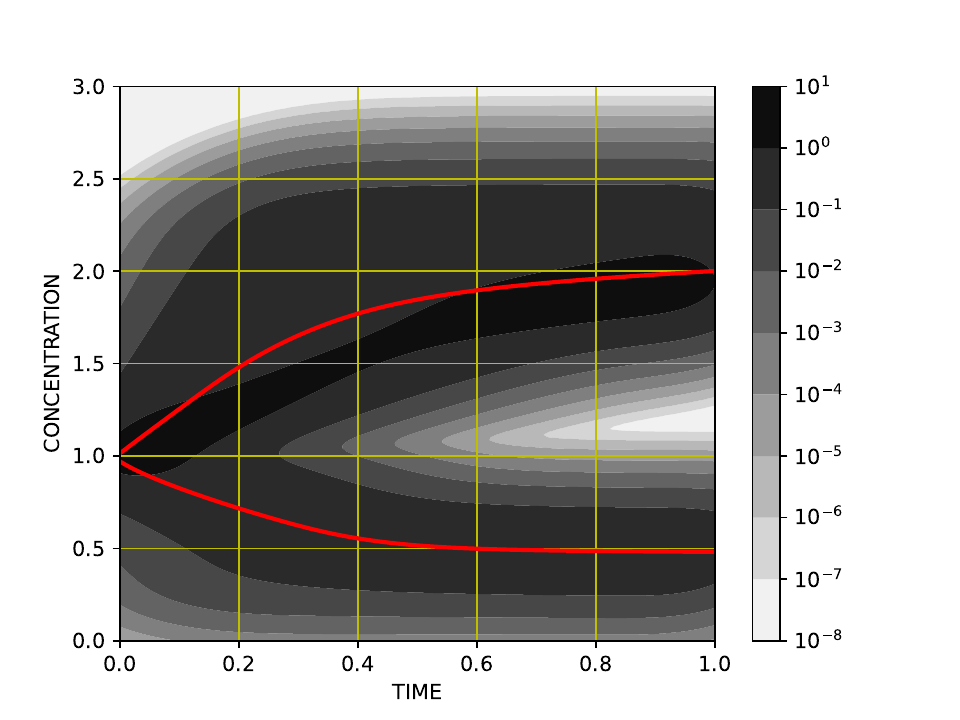}
\caption{LS density}
\end{subfigure}
\begin{subfigure}[b]{0.3\textwidth}
\includegraphics[width=0.99\linewidth]{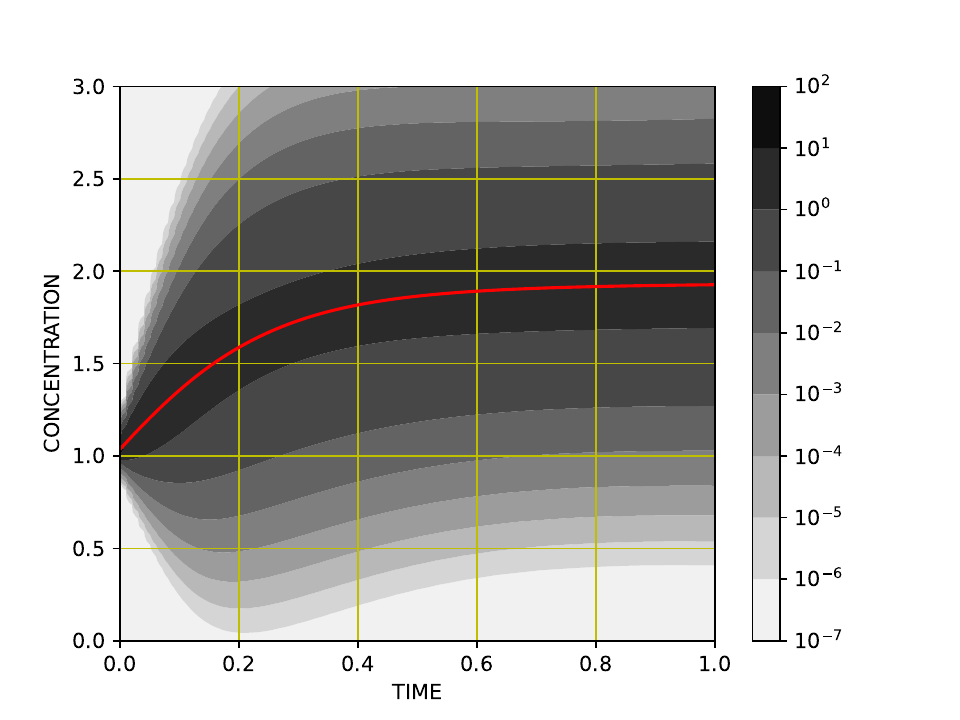}
\caption{BBVI density }
\end{subfigure}
\caption{\Schl \ reaction:
trajectories (upper panels) and density (lower panels).
}
\label{fig:schlogl_trajectories}
\end{figure}

The Schl\"{o}gl system involves  three species $\{ X, A, B \}$ and two reactions
\begin{alignat}{2}
\text{R1}: && A + 2 X & \xrightleftharpoons[k_2]{k_1} 3 X \\
\text{R2}: && B       & \xrightleftharpoons[k_4]{k_3} X
\end{alignat}
The propensities $\{a_i\}$ of these reactions to occur in an increment of (physical) time $\mathrm{d}t$ are
\begin{eqnarray}
\{a_{\rightarrow}, a_{\leftarrow}\}_1 &=& \{k_1 A X (X-1) /2, k_2 X (X-1) (X-2) / 6\} \\
\{a_{\rightarrow}, a_{\leftarrow}\}_2 &=& \{k_3 B, k_4 X\}
\end{eqnarray}
with $k_1 = 3\times 10^{-7}$,  $k_2 = 10^{-4}$, $k_3 = 10^{-3}$, $k_4 = 3.5$.
In the data-generating model, these forward and reverse transitional probabilities drive the Monte Carlo-like stochastic simulation   algorithm (SSA)~\cite{gillespie1977exact,gillespie2007stochastic}.
The output of the SSA for \Schl \ system is multimodal and depends on the particular realization of the stochastic diffusion process and the initial conditions (see \fref{fig:schlogl_trajectories}a).

We choose to model the uncertainty in the model for \Schl \ trajectories with a stochastic framework similar to the one used for the previous exemplar in \eref{eq:langevin_snode}. For the current exemplar we only model and observe the (minor) species $\observation = \hiddenstate = X$.
For this exemplar we considered a model for the right-hand side $\NN_\rhs$ with two internal layers with 16 nodes on each layer and $\tanh$ activations, resulting in 321 weights. The LS employs a drift term $\NN_\weights$ with 2 internal layers with 101 nodes each and $\tanh$ activations.
We used an uninformative prior and one stage of annealing from $\gamma_0=2$ to $\gamma_0=1$ for this demonstration to distribute the replica trajectories across the complex posterior.

\fref{fig:schlogl_trajectories}
compares the data and predicted trajectories via LS as well as Gaussian mixture representations of densities.
After a short-term zone of bifurcation from the initial conditions,
two modes are apparent with roughly equal weight: one with a narrow long time distribution and another at higher values with a broader distribution.
There are small qualitative discrepancies between the predicted trajectories and those of the data generating model most likely due to the (chosen) lack of stochastic component of the data-fit NN model, although the limited representational capacity of the NN and an undersampling of the likely parameters may also play a role.
Nevertheless the LS hypernetwork does capture the the lower, narrow mode and the broader upper mode, while BBVI with a diagonal covariance for efficiency only explores the wider upper mode.

\begin{figure}[htb]
\centering
\includegraphics[width=0.45\linewidth]{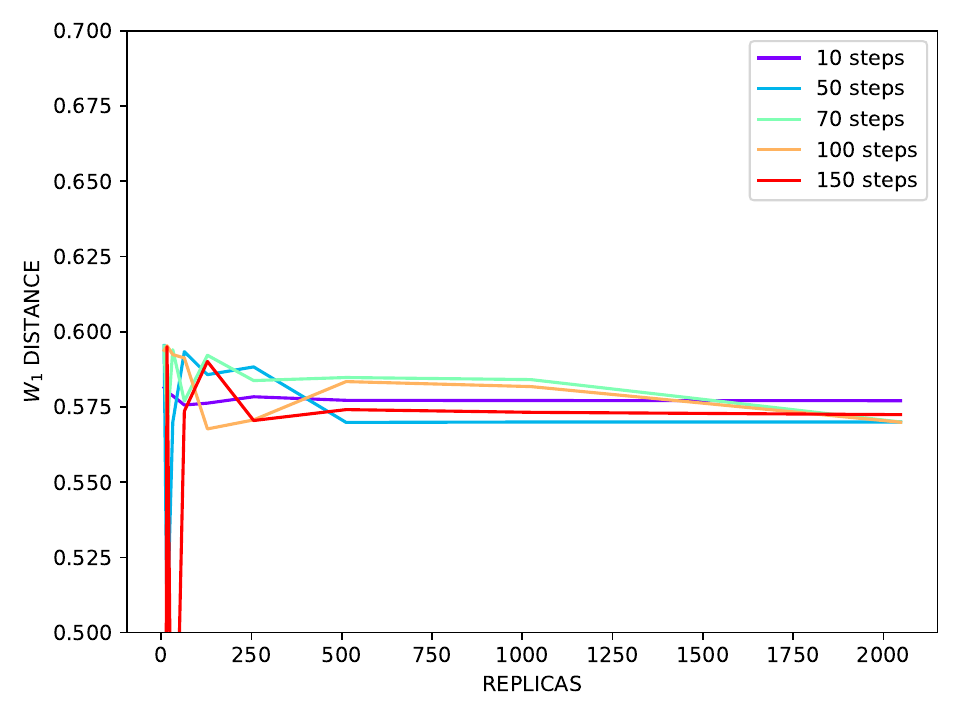}

\caption{\Schl \ reaction:
Wasserstein $\mathcal{W}_1$ distance as function of number of replica samples for sequence of sample pseudo-time steps. }
\label{fig:schlogl_convergence}
\end{figure}

\fref{fig:schlogl_convergence} shows that the convergence of the method in terms of the number of replicas and length of the SDE sampling is both fairly rapid and relatively insensitive to these hyperparameters for this physical system.
There does appear to be a loose trend to converge in terms of replicas faster for longer sampling trajectories.

\begin{figure}[htb]
\centering
\begin{subfigure}[c]{0.4\textwidth}
\includegraphics[width=0.98\textwidth]{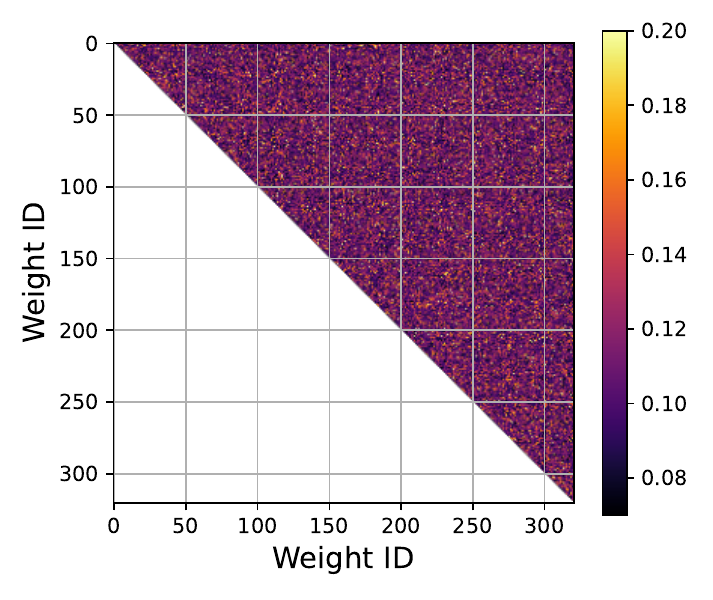}
\caption{Distance Correlation}
\end{subfigure}
\begin{subfigure}[c]{0.4\textwidth}
\includegraphics[width=0.94\textwidth]{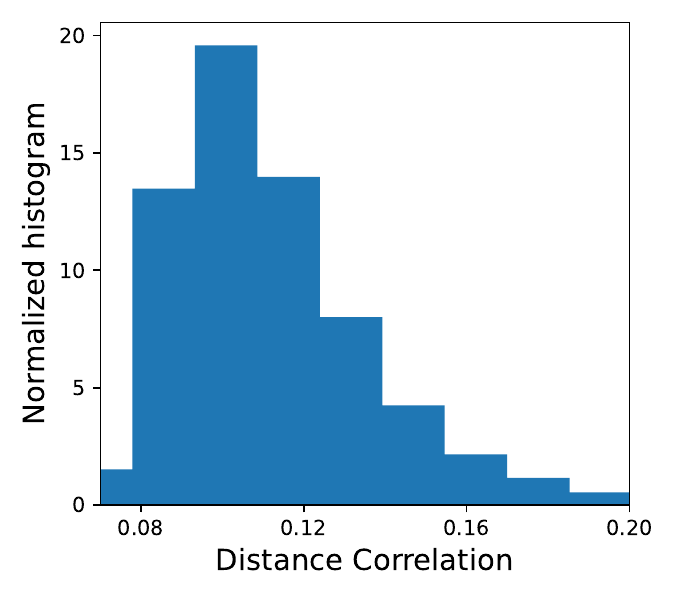}
\caption{Normalized histograms}
\end{subfigure}
\caption{
Distance correlation values for the parameters of (a) $\NN_{\rhs}$ and (b) their normalized histograms corresponding to the \Schl\ model.
\label{fig:schlogl_weights}
}
\label{fig:schlogl_dcorr}
\end{figure}

\fref{fig:schlogl_weights} illustrates the distance correlations between the weight parameters for the $\NN_{\rhs}$ model.
Surprisingly, this measure of dependency only shows weak connections between pairs of model parameters with the bulk of distance correlations hovering around $0.1$.
We conjecture that this is due to the fungibility of network weights which can result in several processing pathways contributing a fraction of the output.

\subsection{Material physics}\label{sec:matphys}

\begin{figure}[!htb]
\centering
\includegraphics[width=0.32\textwidth]{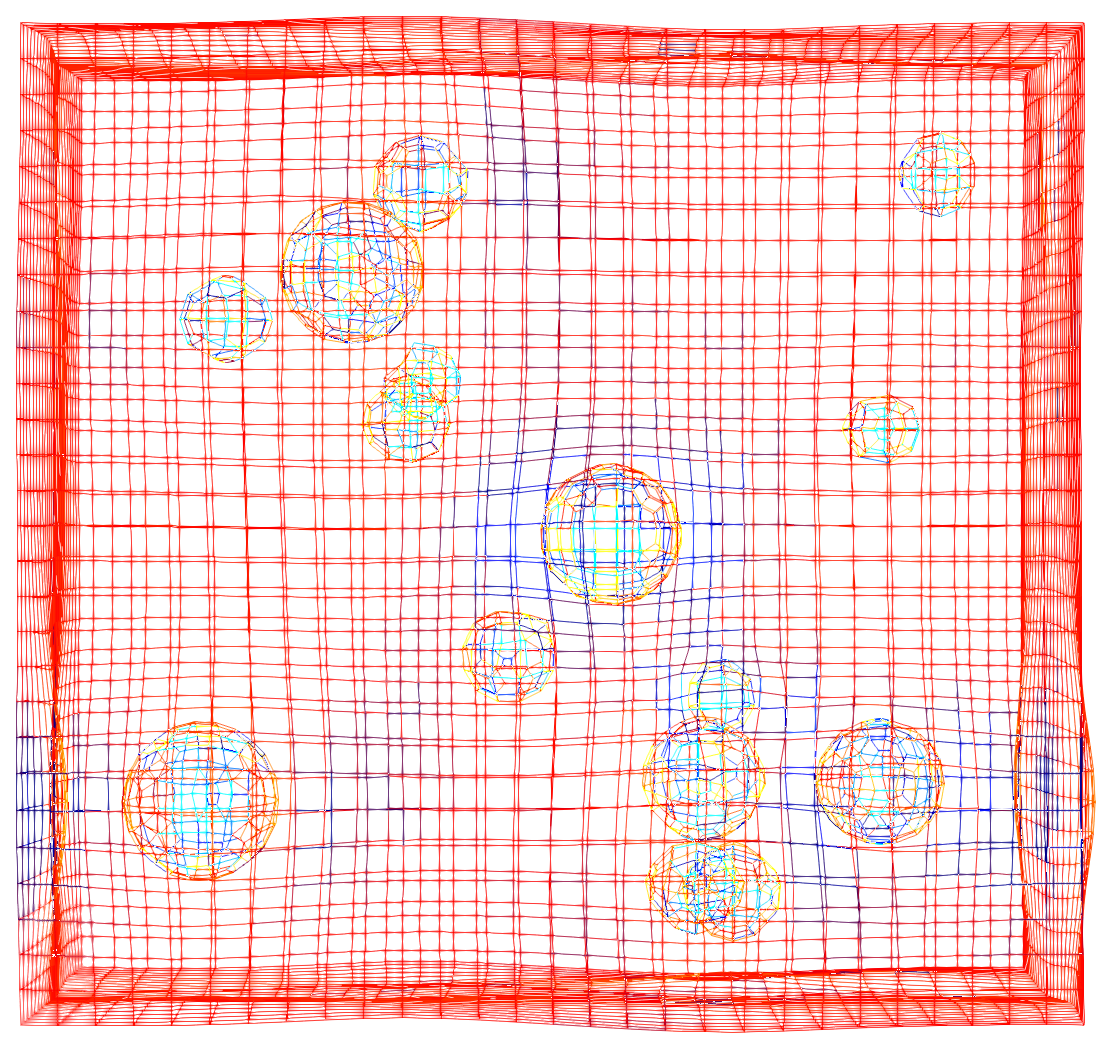}
\includegraphics[width=0.32\textwidth]{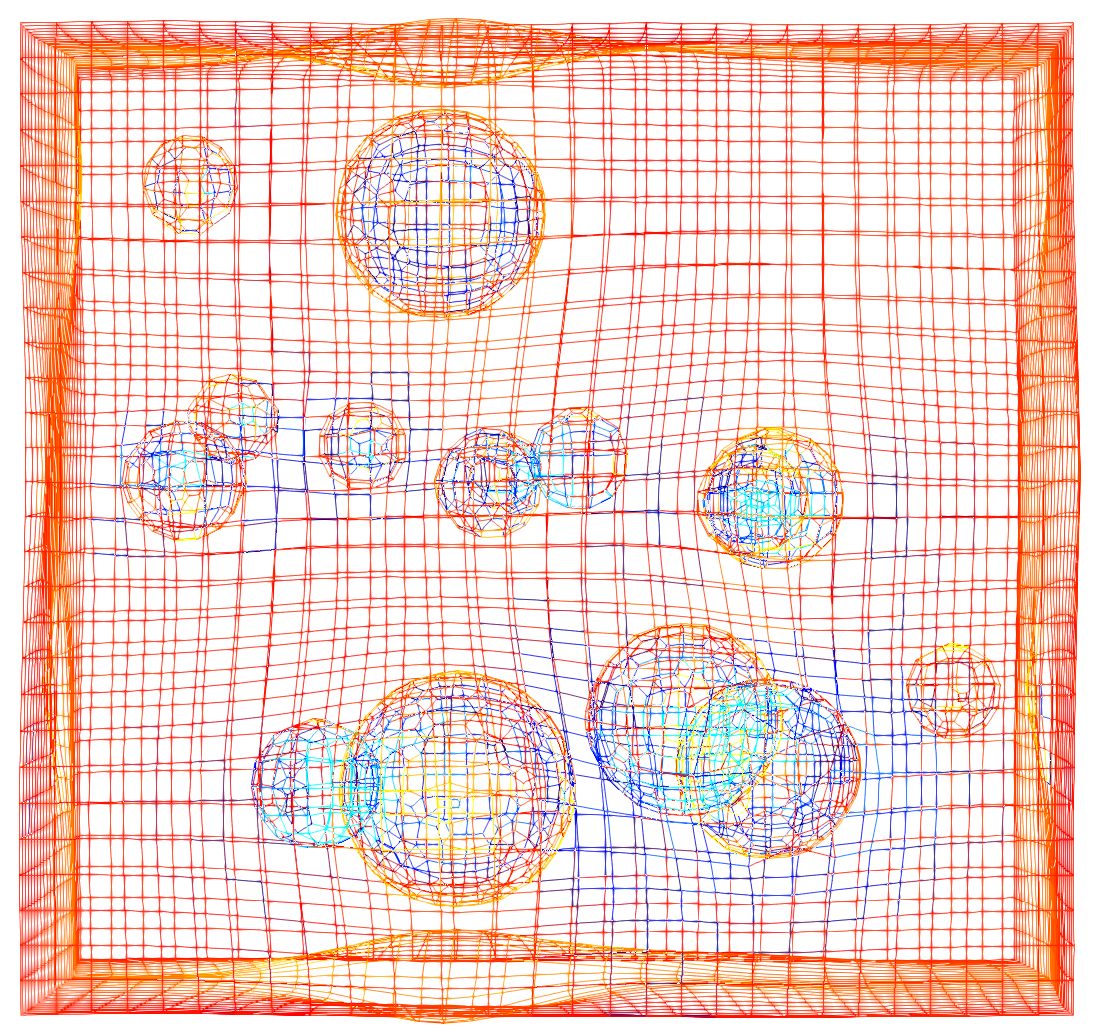}
\includegraphics[width=0.32\textwidth]{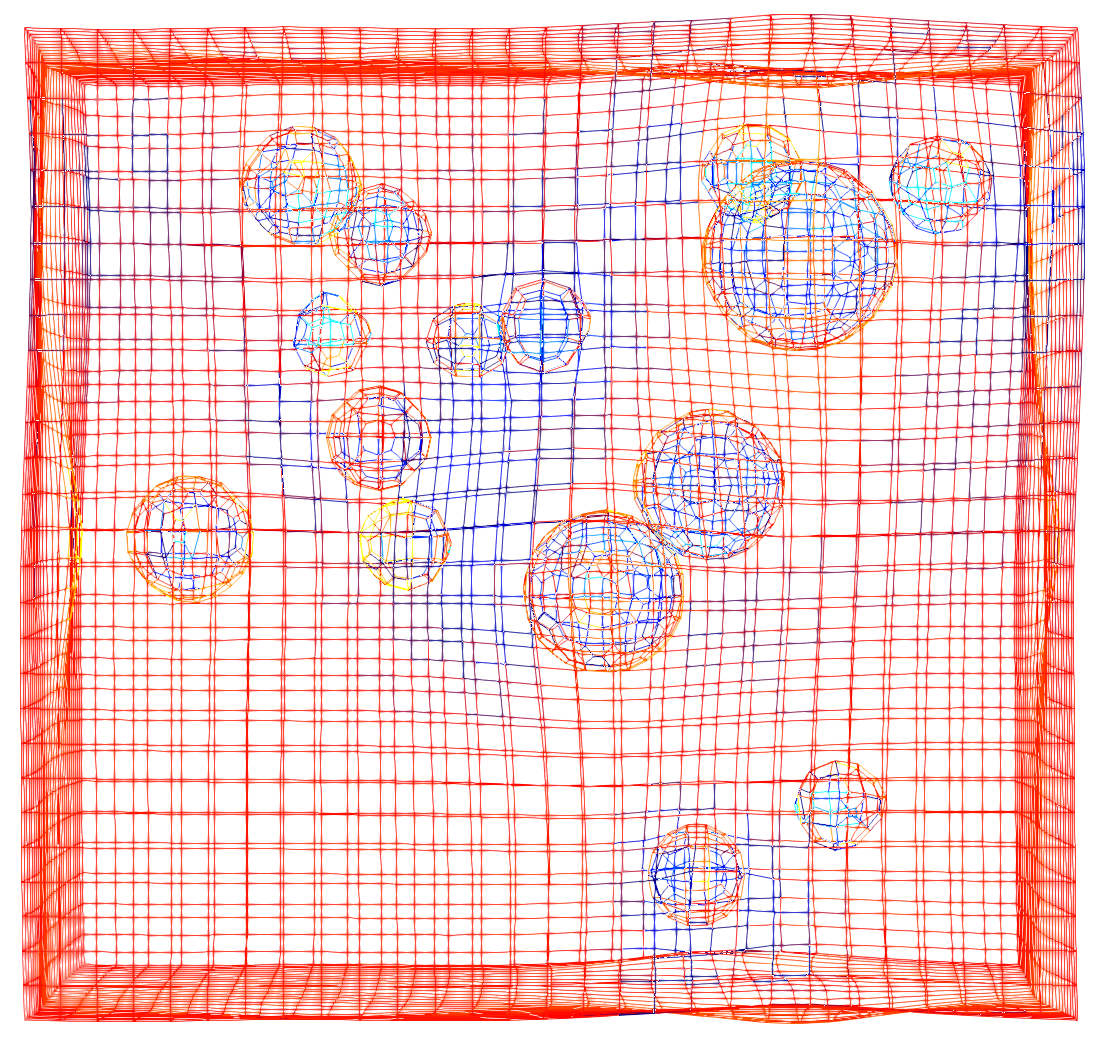}

\caption{Viscoelastic matrix with elastic inclusions. Three realizations in biaxial stretch. Stress is indicated by color and deformation by deviations from a cubic configuration.}
\label{fig:viscoelastic_inclusion_realizations}
\end{figure}

For this exemplar we use the stress response of periodic stochastic volume elements (SVEs) composed of a viscoelastic matrix with spherical elastic inclusions.
\fref{fig:viscoelastic_inclusion_realizations} shows representative realizations.
Here, we employ a simplified version of the simulations in \cref{jones2022neural}.
The stress response of the viscoelastic matrix is modeled with a finite deformation version of linear viscoelasticity with relaxation kernels in a Prony series with relaxation times ranging from 1$\mu$s to 3160s.
For this phase, the instantaneous bulk and shear moduli are 920 MPa and 0.362 MPa, respectively, the equilibrium bulk and shear moduli are 920 MPa and 0.084 MPa, and all other parameters are given in \cref{long2017linear}.
The glass inclusions were modeled with a St. Venant elastic model with  Young’s modulus 60 GPa
and  Poisson’s ratio 0.33.
Full details the data generating model are provided in \cref{jones2022neural}.

We subjected the SVEs to a limited number of  loading modes similar to experiments to create the training and validation data.
The face normal displacements were given by random walks to probe a range of the temporal spectrum of the SVE response.
The combined dataset consists of 100 loadings of 32 configuration replicas drawn from the spatial statistics.
This source of variability represents a source of aleatory uncertainty since the data model, as constructed, is uninformed by the configuration of the replicas.

For modeling this data, we adapted the general data-fitting model from \sref{sec:method} to resemble the ISV-NODE model of \cref{jones2022neural}, which consists of (a) the evolution of internal state $\hiddenstate$
\begin{equation} \label{eq:flow}
\dot{\hiddenstate} = \rhs(\strain, \dot{\strain}, \hiddenstate)
\end{equation}
analogous to \eref{eq:flow}, and the second Piola-Kirchoff stress $\stress$ response from a potential $\energy$
\begin{equation} \label{eq:stress}
\stress = \partialb_\strain \energy(\strain,\hiddenstate)
\end{equation}
which serves as the observation component \eqref{eq:obs_model}.
Using the physical principle of coordinate frame invariance, we compact the potential representation in \eref{eq:stress} using the scalar deformation invariants $\invariants = \{ \tr \strain, \tr \strain^*, \det \strain\}$ of the strain $\strain$ together with the hidden variables $\hiddenstate$, which we assume are also scalar invariants.
Here $\strain^* = \det(\strain) \strain^{-T}$,
and
\begin{equation}
\stress = \partialb_\strain \energy(\invariants, \hiddenstate)
= \partialb_{I_a} \energy(\invariants, \hiddenstate)
\partialb_{I_a} \strain
\end{equation}
where $\{ \partialb_{I_a} \strain \}$ form the natural basis for the stress representation and the partial derivatives of the unknown potential $\partialb_{I_a} \energy$ provide the coefficients of this basis.
The flow depends on scalar invariants of both the strain $\strain$ and its rate $\dot{\strain}$:  $\invariants_{\strain,\dot\strain} = \invariants_{\strain} \cup \{ \tr \dot{\strain}, \tr \dot{\strain}^2, \tr \dot{\strain}^3 \}  \cup \{ \tr \strain \dot{\strain}, \tr \strain^2 \dot{\strain}, \tr \strain \dot{\strain}^2, \tr \strain^2 \dot{\strain}^2 \}$ together with the hidden variables $\hiddenstate$.

We use NNs to represent $\rhs$ and $\energy$
so that the data fitting model is:
\begin{eqnarray}
\dd \hiddenstate &=& \NN_\rhs(\invariants_{\strain,\dot\strain},\hiddenstate; \wH) \dd t \label{eq:flowNN}\\
\observation &=&
\partialb_\Eb
\NN_\observation(\invariants_{\strain} , \hiddenstate; \wO) \label{eq:output_stress}
\end{eqnarray}
where the weights are  $\weights = [ \wH, \wO ]$,
the inputs are  $\inputs = \strain(t)$, and the observable outputs are $\observation = \stress(t)$ .

We employed a 6-dimensional hidden state $\hiddenstate$ that complements a 10-dimensional time-dependent set of invariants $\invariants_{\strain,\dot{\strain}}$ for a total of 16 input values.

\subsubsection{Hyperelastic potential}

In a limited regime, the hyperelastic material behaves elastically such that the internal state does not change $\dot{\hiddenstate} = \mathbf{0}$, i.e. $\NN_{\rhs}\equiv 0$.
For this case we collect data at high rates and only train the observation model that relates input strain to observable stress.

Given the hidden state is not used for this exemplar, the input state for $\NN_\observation(\invariants_{\strain}; \wO) $ consists of 3 invariants $\invariants_{\strain}$, and the model uses 2 hidden layers with the same number of neurons as the input state, {\it softplus} activations, and a linear last layer for a total of $\vert\weights\vert=28$ weights.
We present three test cases for this model, all with the same configuration for the sampler drift $\NN_{\weights}$: 2 hidden layers with the same number of neurons as the number of inputs space, {\it tanh} activations, and a linear last layer.
In the first case, \emph{full} UQ, we cast all $\wO$ as uncertain, resulting in $\vert\parameters\vert=2436$ parameters for $\NN_{\weights}$.
For the second case, \emph{BLL+fixed $\wO^{(d)}$}, only the parameters of the last layer (three weights and one bias) are considered uncertain, denoted as $\wO^{(s)}$.
The remaining set of parameters $\wO^{(d)}$, corresponding to the hidden layers, are fixed at their maximum likelihood estimate (MLE).
For this case, $\vert\parameters\vert=60$.
Finally, the third case, \emph{BLL+reoptimized $\wO^{(d)}$}, is similar to the second case except we relax the MLE constraint for $\wO^{(d)}$, and these parameters are optimized jointly with the parameters of the score function $\NN_\rhs$.
For this case, the total number of parameters optimized by the proposed framework is $\vert\wO^{(d)}\vert+\vert\parameters\vert=24+60=84$.
We refer to cases, BLL+fixed and BLL+reoptimized, when the uncertainty is associated with the last layer only, i.e. Bayesian last layer (BLL)~\cite{lazaro2010marginalized,watson2021latent}.

\begin{figure}[htb]
\centering
\begin{subfigure}[c]{0.4\linewidth}
\includegraphics[width=0.99\linewidth]{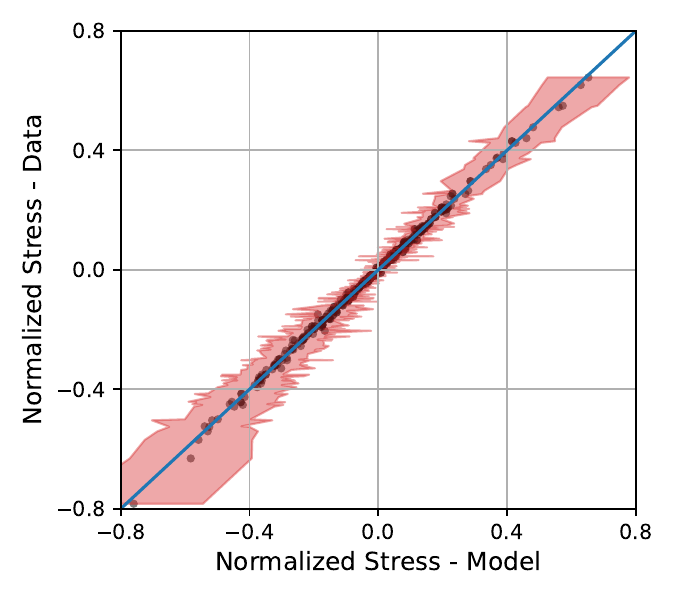}
\caption{training}
\end{subfigure}
\begin{subfigure}[c]{0.4\linewidth}
\includegraphics[width=0.99\linewidth]{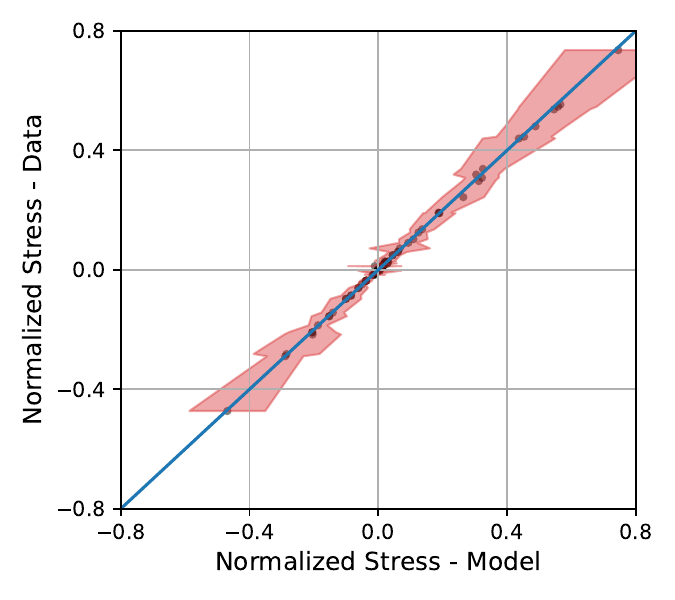}
\caption{validation}
\end{subfigure}
\caption{Hyperelastic data: parity plot for a MLE model fit  (a) training data, (b) validation data. Black circles show mean data values plotted against the MLE model solution. The red shaded area shows $\pm$ one standard deviation around the data mean and is thus represented horizonthaly.}
\label{fig:elastic_accuracy_map}
\end{figure}

We start with the MLE fit for the hyperelastic potential.
\fref{fig:elastic_accuracy_map} shows parity plots for both the training data and the validation data that was not seen during the training process.
We observe some data fit discrepancies but these discrepancies are generally on par with data variance, the extent of which is shown in red, for both the training and the validation data.
Note that the range of extent of the training data and validation data differs due to the random approach used to split the available data into training, test, and validation subsets.
This MLE fit was then used as the initial condition for the weights in the Langevin sampling SDE.

\begin{figure}[htb]
\centering
\includegraphics[width=0.9\linewidth]{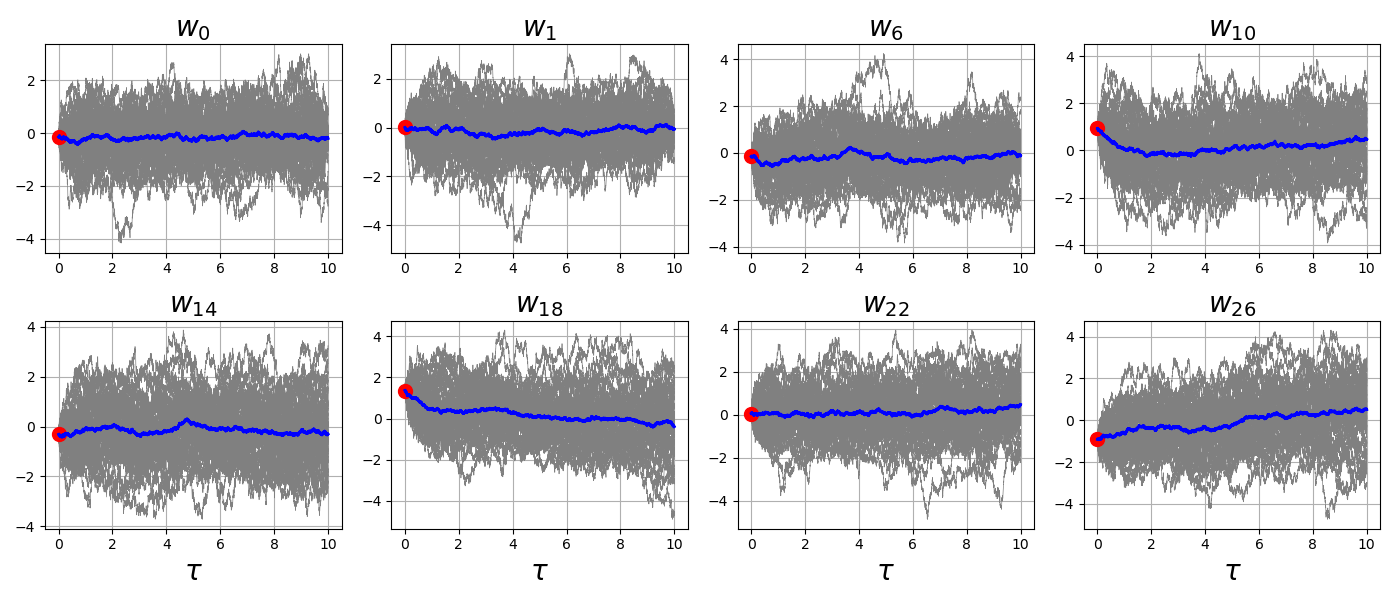}
\caption{Hyperelastic: SDE trajectories for select parameters of $\NN_{\energy}$ in pseudo time $\tau$.
Trajectories start from initial values that are randomly perturbed conditions around the MLE. The mean is shown in blue, and the MLE is shown with a red symbol.}
\label{fig:elastic_weights}
\end{figure}

\begin{figure}[htb]
\centering
\begin{subfigure}[c]{0.32\linewidth}
\includegraphics[width=0.99\linewidth]{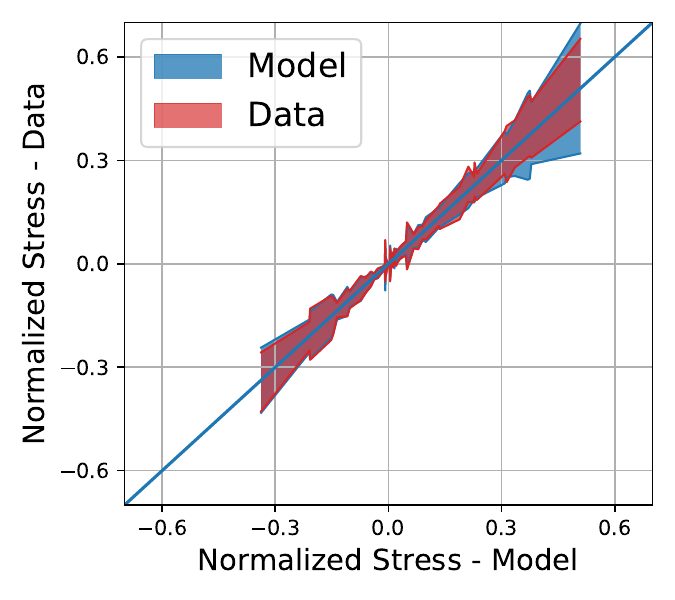}
\caption{Full}
\end{subfigure}
\begin{subfigure}[c]{0.32\linewidth}
\includegraphics[width=0.99\linewidth]{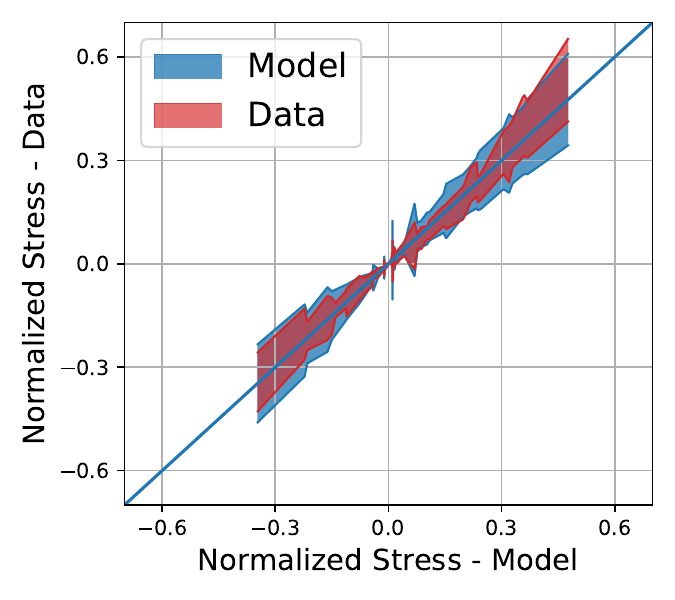}
\caption{BLL + fixed $\wO^{(d,\text{MLE})}$}
\end{subfigure}
\begin{subfigure}[c]{0.32\linewidth}
\includegraphics[width=0.99\linewidth]{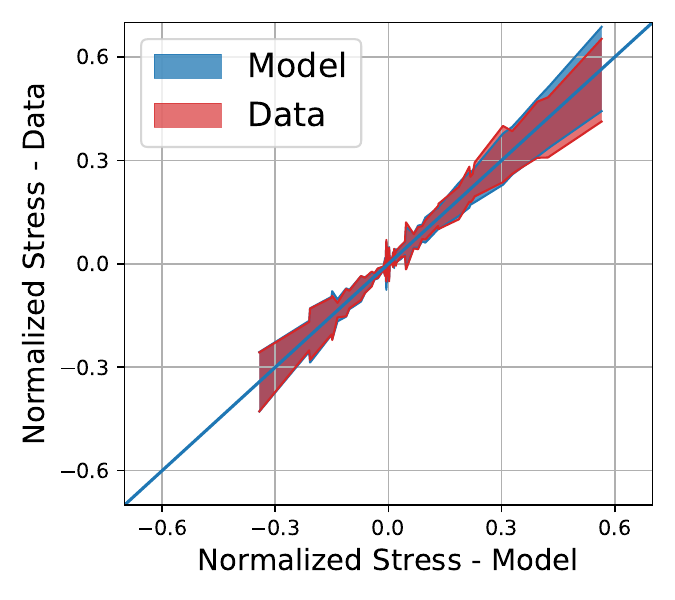}
\caption{BLL + re-optimized $\wO^{(d)}$}
\end{subfigure}
\caption{Hyperelastic data: parity plot for a stochastic model fit of the validation data. The shaded area shows the mean $\pm 1$ standard deviation in the data and black symbols represent model realizations plotted against the data. The blue and black symbols show the standard deviation values for the data and the model. The pseudo-time step is $\Delta\tau = 10^{-3}$.}
\label{fig:elastic_accuracy}
\end{figure}

\fref{fig:elastic_weights} shows randomly selected weight trajectories from both layers of the data model.
Clearly there is good mixing with some evidence of transient decay from initial MLE weights, e.g. for $\weights_{10}$.
The blue lines, representing means over all trajectories for each weight, exhibit some noise due to the finite sample size (50 samples in this case), however the results are sufficient to gauge the required number of SDE steps (in pseudo-time $\tau$) to ensure the Langevin samples are collected after the completion of any transient period.
We employed between $5\times 10^3$ and $2\times 10^4$ Langevin steps for all models presented in this section.
Note the O(1) variance in the NN weights maps to much smaller relative variance in\ the outputs, as shown in \fref{fig:elastic_accuracy}.

Lastly, \fref{fig:elastic_accuracy} compares the three cases: (a) full where all model parameters are stochastic, (b) BLL with other parameters fixed at their MLE values, (c) BLL with other parameters re-optimized using the MLE as initial conditions.
It appears that (b) slightly underestimates the uncertainty in the outputs, while (c) provides uncertainty estimates on par and perhaps slightly better compared to the full treatment (a) at a significant savings in sampling costs.
The slightly improved converage of the data uncertainty by (c) compared to (a) is likely due to improved sampling efficiency given a lower dimensional stochastic space, 4 compared to 28.
In fact the $\Wc_1$ distance for the BLL+reoptimize variant (c) is 0.71 of the $\Wc_1$ distance for the full treatment (a), while ratio is  1.18 for the BLL+fixed variant (b).

\subsubsection{Viscoelastic composite material}

In this case we train the full data model i.e. both the flow of internal states, in \eref{eq:flowNN}, and the stress observation model, in \eref{eq:output_stress}.
Based on the fact that the internal states and their flow are generally more uncertain than stress-strain relations, we explore simplifications of the UQ procedure where only subsets of the model parameters are explored with Langevin sampling.

The $\NN_{\rhs}$ component had 2 hidden layers with the same number of neurons as the input  state and each layer employs a $\tanh$ activation function.
A final linear layer then outputs the time derivative of the 6-dimensional hidden state, resulting in a total of $646$ parameters for $\NN_{\rhs}$.
The observation model $\NN_{\observation}$ employs a 9-dimensional input consisting of 3 invariants $\invariants_{\strain}$ and 6 hidden variables.
This feedforward $\NN$ had 2 hidden layers, both with the same number of neurons as the input size and {\it softplus} activation function.
The last, linear layer outputs a scalar potential, which then produces the stress tensor via \eref{eq:output_stress}.
The number of parameters is $561$ for $\NN_{\observation}$.
Thus the total number of parameters for this data model is $1207$.

\begin{figure}[!htb]
\centering
\begin{subfigure}[c]{0.4\textwidth}
\includegraphics[width=0.99\textwidth]{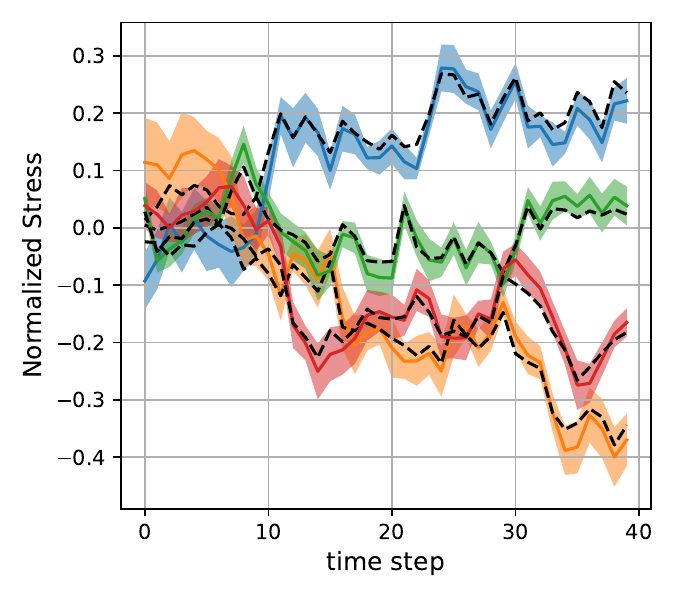}
\caption{selected sample trajectories}
\end{subfigure}
\begin{subfigure}[c]{0.4\textwidth}
\includegraphics[width=0.99\textwidth]{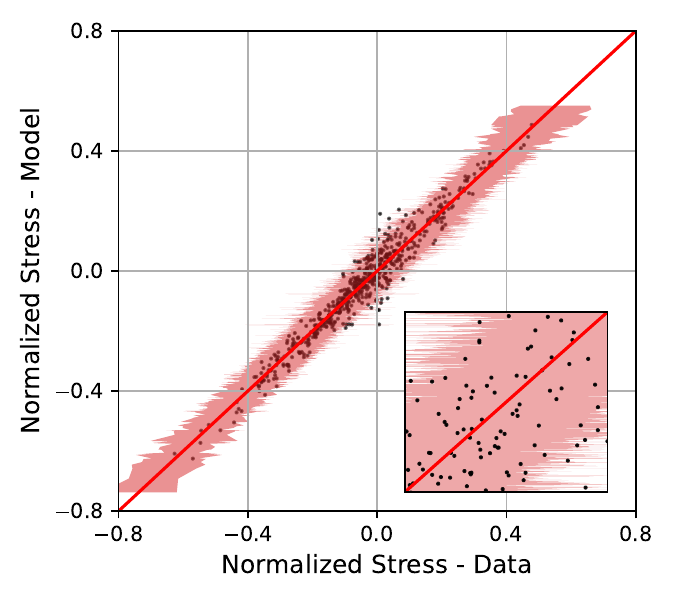}
\caption{parity plot over all samples}
\end{subfigure}
\caption{Viscoelastic composite data: MLE fit of a NODE model through the viscoelastic data held out for validation. (a) select data trajectories in color and corresponding MLE predictions in dashed black lines. (b) parity plot with model predictions as black dots and data variance as blue band around red parity line.}
\label{fig:viscoelastic_inclusion_comparison}
\end{figure}

\fref{fig:viscoelastic_inclusion_comparison} shows a comparison between viscoelastic data and the corresponding NODE model fitted using a MLE cost function~\eqref{eq:likelihood}.
\fref{fig:viscoelastic_inclusion_comparison}a shows the mean normalized stress with thick lines overlaid over shaded areas representing $\pm$ one standard deviation for select simulations trajectories.
The dashed black lines represent the model predictions corresponding to the MLE solution.
The black symbols in the parity plot represent MLE model predictions collected over all trajectories and stress tensor components and are similarly laid over a shaded region representing two standard deviations around the mean validation data.
While it would appear the model lacks the capacity to closely capture the data trends at small stress values, the inset in \fref{fig:viscoelastic_inclusion_comparison}b, representing a magnified view near the range $[0,0.1]$ shows large variations in how the data is spread around the mean.
The model samples in this region fall within two standard deviations around the local mean and thus the MLE estimate weighs these samples similarly with other model predictions that correspond to data exhibiting lower uncertainties.

Next, we show results obtained via the proposed hypernetwork.
For this set of results, we utilized both components of the data-fitting neural network model: $\NN_{\rhs}$ and $\NN_{\observation}$.
These models were cast in a variant of the {\it BLL+re-optimized $\weights^{(d)}$} explored in the previous section.
The parameters for the last, linear, layers for both $\NN_{\rhs}$ and $\NN_{\observation}$ were considered stochastic, with values for these parameters generated via Langevin sampling.
These sets are designated by $\wH^{(s)}$ and $\wO^{(s)}$, respectively in~\Aref{alg:training}.
The corresponding model for the drift term, equivalent with the score function, $\NN_{\weights}$ thus consists of $119$ inputs (given that $\vert\wH^{(s)}\vert=102$ and $\vert\wO^{(s)}\vert=17$), two hidden layers with the same number of neurons as the input state and a {\it tanh} activation function, and a linear last layer for a total of around $43\times 10^3$ parameters.
The total number of parameters that are optimized is thus approximately $44\times 10^3$ when we include the non-stochastic components $\wH^{(d)}$ and $\wO^{(d)}$ corresponding to $\NN_{\rhs}$ and $\NN_{\observation}$, respectively.

\begin{figure}[!htb]
\centering
\includegraphics[width=0.33\textwidth]{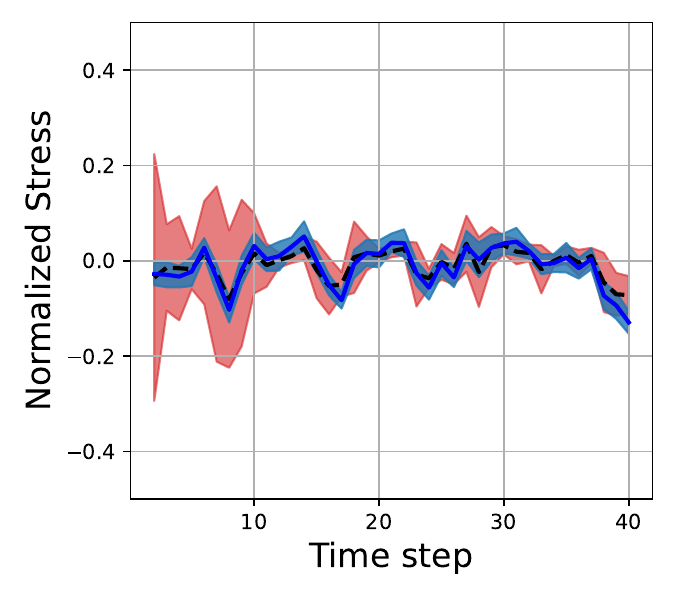}
\includegraphics[width=0.33\textwidth]{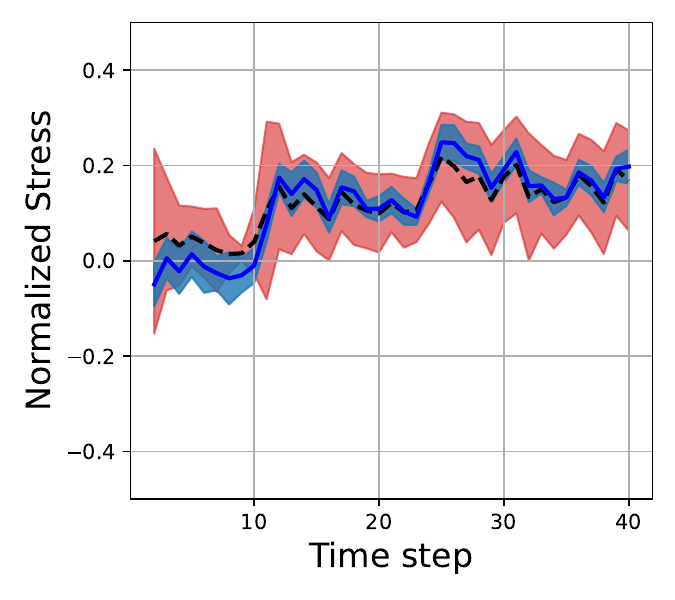}
\includegraphics[width=0.33\textwidth]{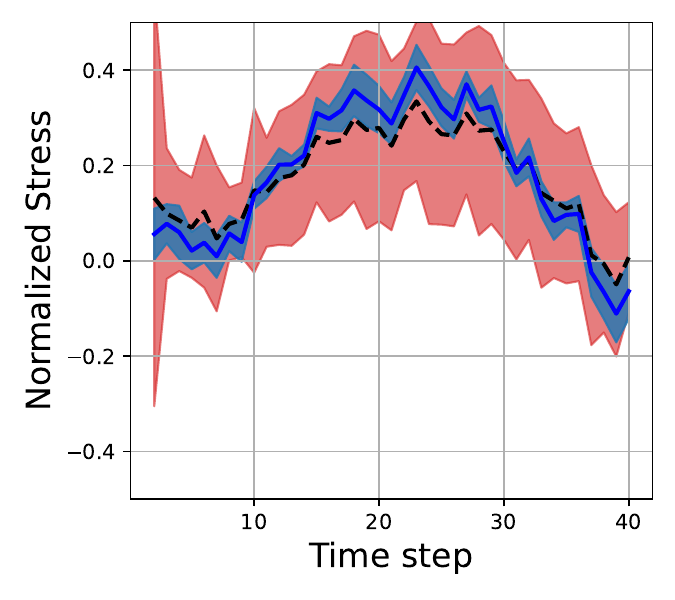}
\caption{Viscoelastic composite response results for selected held-out validation random walk trajectories.
The shaded red region shows one standard deviation around the predicted mean trajectory shown with a dashed black line.
The shaded blue region shows one standard deviation around the mean data trajectory shown with a red line.
}
\label{fig:viscoelastic_samples}
\end{figure}

\fref{fig:viscoelastic_samples} compares select sample trajectories in the validation set with simulation ensembles via the proposed hybrid calibration framework. These represent conditions for the model and data uncertainty are on par (left frame), about the average over all validation data (middle frame), and a case (in the right frame) where the model uncertainty is the largest relatively to the corresponding data.
The dynamics of the mean in the data, shown with blue lines, are captured by the surrogate model framework, shown with dashed black lines.
The uncertainty band gleaned from the surrogate simulation, shown in light red, is larger than the data uncertatinty, shown in blue, in particular when the model predictions are shifted away from the mean trends.
The mean standard deviation associated with the stochastic model predictions is 2-4 times larger compared to the mean standard deviation for each sample.
This is not unexpected given the BLL variational inference framework employed for computational efficiency.

\begin{figure}[htb]
\centering
\begin{subfigure}[c]{0.32\textwidth}
\includegraphics[width=0.98\textwidth]{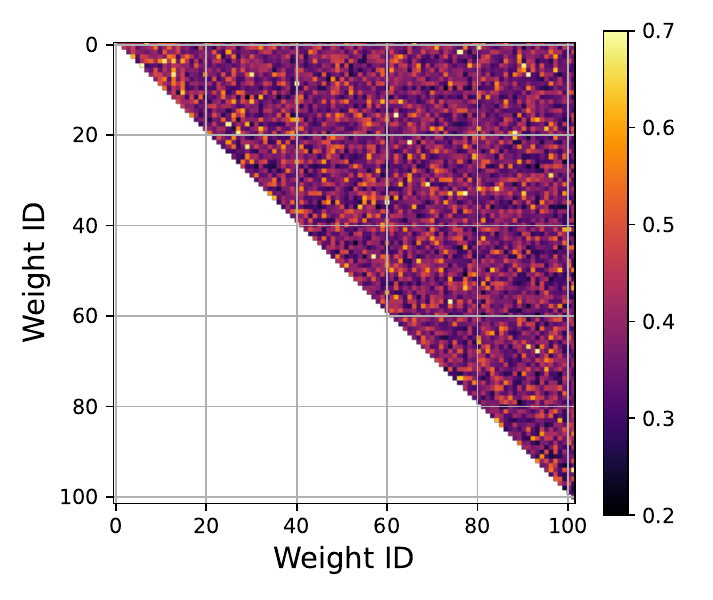}
\caption{Hidden State Model - $\NN_{\rhs}$}
\end{subfigure}
\begin{subfigure}[c]{0.32\textwidth}
\includegraphics[width=0.98\textwidth]{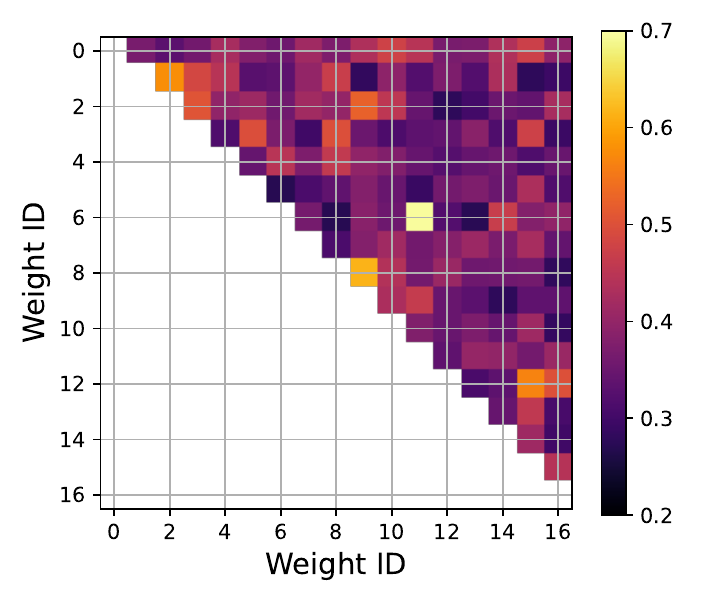}
\caption{Observation Model - $\NN_{\observation}$}
\end{subfigure}
\begin{subfigure}[c]{0.29\textwidth}
\includegraphics[width=0.98\textwidth]{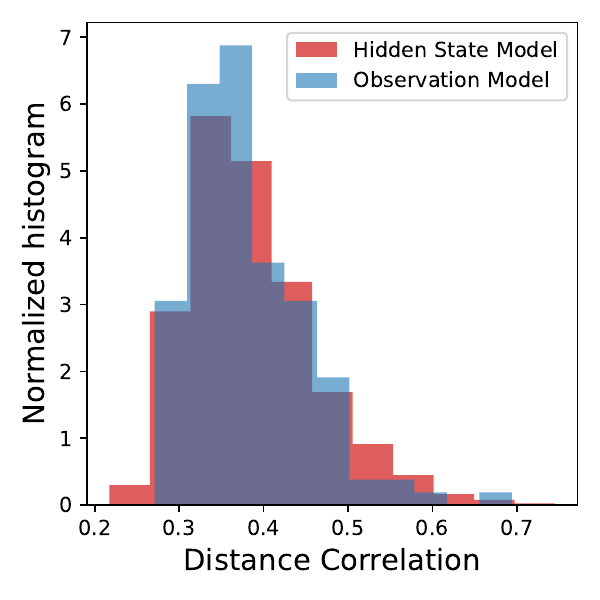}
\caption{Normalized histograms}
\end{subfigure}
\caption{
Distance correlation values for the parameters of (a) $\NN_{\rhs}$ and (b) $\NN_{\observation}$ and (c) a comparison of their distributions vianormalized histograms.
}
\label{fig:viscoelastic_dcorr}
\end{figure}

Lastly, we employ distance correlation~\cite{szekely2007measuring,szekely2009brownian} to estimate the degree of dependence between the inferred model parameters.
\fref{fig:viscoelastic_dcorr} shows distance correlation statistics, computed using $50$ surrogate model replicas, for $\wH^{(s)}$ and $\wO^{(s)}$ associated with the hidden state model and observation model, respectively.
Recall that only the parameters corresponding to the last layers of these models were assumed to control the uncertainty in the predictions of these model while all the other parameters follow a classical optimization loop.
These results indicate a non-negligible degree of dependence between parameters of either model.
The histogram in \fref{fig:viscoelastic_dcorr}c shows a similar distribution for this quantity for both the hidden state model and the observation model.
We then constructed the distance correlation between the ensemble of parameters for the hidden state model and the ones for the observation model is approximately $0.95$.
This suggests that while individual parameters weakly compensate for other parameters of the same model due to redundancy, overall the dependence between the hidden state model and the observation model is quite strong.

\section{Conclusion} \label{sec:conclusion}

This work presents a hierarchical framework that combines sampling methods with optimization to endow large-scale models with the capability to generate predictions commensurate with the discrepancy between the model and available training data.
The methodology relies on a hypernetwork the utilizes Langevin sampling to generate ensembles of model parameters in a variational inference context.
The framework simultaneously optimizes a score function that characterizes the posterior distribution for a subset of the physical model parameters while directly searching for optimal values for the other model parameters.
As with other UQ methods, samples of the resulting parameter distributions can be pushed forward to provide epistemic (due lack of training data) and aleatory (due to inherent variability) UQ on quantities of interest for engineering design and performance evaluation.
The method is also adaptable to generating likely responses on-the-fly so that these could be embedded in a suitable simulation host code or optimal experimental design method \cite{fedorov2010optimal}.

We demonstrated the algorithm on canonical chemical reaction exemplars and also presented its application to complex material physics.
For the chemical reaction cases we compared our predictions with standard mean-field variational inference results.
These comparisons show that the proposed LS hypernetwork  is better equipped to capture the uncertainty in the training data compared to standard variational inference models.
Using the material physics exemplars, we showed that layer or component-level stochastic parameters, akin to Bayesian last layer algorithms albeit with reoptimized point estimates, generate predictions similar to the more expensive fully stochastic ensembles that include all model parameters.
The proposed framework is applicable to a wide variety of heterogeneous complex models and provides a competing alternative to ensemble methods such as the Stein variational gradient descent algorithm that was recently coupled with sparsification methods by our group~\cite{padmanabha2024improving,padmanabha2024condensed}.

In following work, we plan to employ the  encoder-decoder U-Nets~\cite{ronneberger2015u}, widely used in diffusion models~\cite{si2024freeu}, to model the score function.
This will also provide latent space for the score that may be useful in providing more interpretability for UQ studies.
We will also pursue methods to embed physical constraints on the posterior, such as weight bounds and specialized NNs such as \emph{input convex} NNs~\cite{amos2017input}.
We may also explore alternatives to Langevin sampling in the overall hypernetwork and seek methods that coordinate the Langevin samples~\cite{garbuno2020interacting}.

\section*{Acknowledgments}

We employed JAX \cite{jax2018github} and Equinox \cite{kidger2021equinox} in the development of this work, which we gratefully acknowledge.

This material is based upon work supported by the U.S. Department of Energy, Office of Science, Advanced Scientific Computing Research program.
The authors were supported by Advanced Scientific Computing program at Sandia National Laboratories.
CS was also supported by the Scientific Discovery through the Advanced Computing (SciDAC) program through the FASTMath Institute.
Sandia National Laboratories is a multimission laboratory managed and operated by National Technology and Engineering Solutions of Sandia, LLC., a wholly owned subsidiary of Honeywell International, Inc., for the U.S.
Department of Energy's National Nuclear Security Administration under contract DE-NA-0003525.
This paper describes objective technical results and analysis.
Any subjective views or opinions that might be expressed in the paper do not necessarily represent the views of the U.S.  Department of Energy or the United States Government.


\appendix
\raggedbottom
\setcounter{equation}{0}
\renewcommand{\theequation}{A-\arabic{equation}}
`\section{The Kullback-Liebler divergence between two random processes} \label{app:KLD}

Following \crefs{opper2019variational,archambeau2007gaussian,archambeau2007variational} we provide a more step-by-step derivation of the joint Kullback-Liebler (KL) divergence between two processes associated with SDEs and a bound on the KL divergence between corresponding steps of the two processes.

Consider an Euler-Maruyama approximation of two SDEs with the same diffusion term
\begin{eqnarray}
\xs_{k+1}&=&\xs_k+\fs(\xs_k)\Delta{\tau}+\sigma {\Delta}\Bs_k\\
\xs_{k+1}&=&\xs_k+\gs(\xs_k)\Delta{\tau}+\sigma {\Delta}\Bs_k
\end{eqnarray}
where $\Bs$ is Brownian motion (a Wiener process) and $\Delta{\Bs_k}\propto \Nc(\mathbf{0},\sqrt{\Delta{\tau}}\Is)$.
These SDEs generate a stochastic process, $\{\xs_0,\xs_1,\hdots,\xs_n\}$, over times $\{\tau_0,\tau_1,\hdots,\tau_n\}$, which we associate with the distributions $q$ and $p$, respectively.
In this appendix we will derive the KL divergence between these two processes, both in terms of sequence-to-sequence and sample-to-sample divergences.

We will use the notation,
\begin{eqnarray}
q_k     &\equiv& q(\xs_k) \\
q_{k|j} &\equiv& q(\xs_k|\xs_j) \\
q_{[i,j]} &\equiv& q(\xs_i,\xs_{i+1},\hdots,\xs_j) \\
q_{[i,j]|[k,l]} &\equiv& q(\xs_i,\hdots,\xs_j|\xs_{k},\hdots,\xs_l) \\
d\xs_{[0,n]} &\equiv& \prod_{k=0}^{n}d\xs_k
\end{eqnarray}
to simply the following developments
We assume the Markov property, i.e., $q_{i+1|[0,i]} = q_{i+1|i}$;  as a result, the distribution of the path is:
\begin{equation}
q_{[0,n]}
= q_0 q_{1|0} q_{2|1} \hdots q_{n|n-1}
= q_0 \prod_{k = 0}^{n-1} q_{k+1|k}
= q_0 q_{[1,n]|0}
\end{equation}
where $q_0$ is the distribution of the initial state.
The distribution for $p_{[0,n]}$ follows identically.
Note the following identities,
\begin{eqnarray}
\int d\xs_{[i,j]} q_{[i,j]|i-1} &=& 1 \\
\ln q_0 q_{[1,n]|0} &=& \ln q_0 + \sum_{l=0}^{n-1} \ln q_{l+1|l}\\
\int d\xs_{[k,l]} q_{[i,j]|i-1} f(\xs_0,\hdots,\xs_n) &=& q_{[i,j]|i-1} \int d\xs_{[k,l]} f(\xs_0,\hdots,\xs_n) \quad \mathrm{if}\ j<k.
\end{eqnarray}

The KL divergence between the sequences that $q$ and $p$ generate is,
\begin{equation}
\begin{aligned}
&\KL(q_{[0,n]}||p_{[0,n]}) = \KL(q_0 q_{[1,n]|0}||p_0 p_{[1,n]|0}) \\
&= \int d\xs_{[0,n]} q_0 q_{[1,n]|0} \left[ \ln \left( q_0 q_{[1,n]|0} \right) - \ln \left( p_0 p_{[1,n]|0} \right) \right] \\
&= \int d\xs_{[0,n]} q_0 q_{[1,n]|0} \left(\left(\ln q_0 -\ln p_0\right) + \sum_{l=0}^{n-1} \left( q_{l+1|l} - p_{l+1|l} \right)\right) \\
&= \int d\xs_{[0,n]} q_0\left(\ln q_0 -\ln p_0\right) q_{[1,n]|0} +  \int d\xs_{[0,n]} q_0 q_{[1,n]|0} \sum_{l=0}^{n-1} \left( q_{l+1|l} - p_{l+1|l} \right) \\
&= \int d\xs_0 q_0\left(\ln q_0 -\ln p_0\right) \overbracket{\int d\xs_{[1,n]} q_{[1,n]|0}}^{=1} +  \int d\xs_{[0,n]} q_{[0,n]} \sum_{l=0}^{n-1} \left( q_{l+1|l} - p_{l+1|l} \right) \\
&= \KL(q_0||p_0) + \sum_{l=0}^{n-1} \int d\xs_{[0,n]} q_{[0,n]} \left( q_{l+1|l} - p_{l+1|l} \right) \\
&= \KL(q_0||p_0) + \sum_{l=0}^{n-1} \int d\xs_{[0,l]} \int d\xs_{l+1} \int d\xs_{[l+2,n]} q_{[0,l]} q_{l+1|l} q_{[l+2,n]|l+1} \left( q_{l+1|l} - p_{l+1|l} \right) \\
&= \KL(q_0||p_0) + \sum_{l=0}^{n-1} \int d\xs_{[0,l]} q_{[0,l]} \int d\xs_{l+1} q_{l+1|l} \left( q_{l+1|l} - p_{l+1|l} \right) \overbracket{\int d\xs_{[l+2,n]} q_{[l+2,n]|l+1}}^{=1} \\
&= \KL(q_0||p_0) + \sum_{l=0}^{n-1} \int d\xs_{[0,l]} q_{[0,l]} \KL\left( q_{l+1|l} || p_{l+1|l} \right) \\
&= \KL(q_0||p_0) + \sum_{l=0}^{n-1} \Exp_{q_{[0,l]}} \left[\KL\left( q_{l+1|l} || p_{l+1|l} \right) \right].
\end{aligned}
\end{equation}

Using the transition probabilities,
\begin{equation}
\begin{aligned}
q_{l+1|l} = \Nc(\xs_l+f(\xs_l,\tau_l)\Delta \tau,\Sigmab \Delta \tau)\\
p_{l+1|l} = \Nc(\xs_l+g(\xs_l,\tau_l)\Delta \tau,\Sigmab \Delta \tau)
\end{aligned}
\end{equation}
the KL divergence for the transitions is,
\begin{equation}
\begin{aligned}
\KL(q_{l+1|l}||p_{l+1|l})  = \frac{1}{2} \left( f(\xs_l,t_l) - g(\xs_l,t_l)  \right)^T \Sigmab^{-1} \left( f(\xs_l,t_l) - g(\xs_l,t_l)  \right) \Delta t \\
\equiv ||f(\xs_l,t_l) - g(\xs_l,t_l)||_\Sigma^2 \Delta t.
\end{aligned}
\end{equation}
The KL divergence for the paths becomes,
\begin{equation}
\begin{aligned}
\KL(q_{[0,n]}||p_{[0,n]}) &= \KL(q_0 || p_0) + \sum_{l=0}^{n-1} \Exp_{q_{[0,l]}} \left[ ||f(\xs_l,t_l) - g(\xs_l,t_l)||_\Sigma^2  \right]\Delta t.
\end{aligned}
\end{equation}
In the limit, $n \rightarrow \infty$, we obtain a Riemann integral
\begin{equation}
\KL(q||p) = \KL(q_0 || p_0) + \int_0^{\tau_f} \mathrm{d}\tau \,  \Exp_{q} \left[ ||f(\xs(\tau),\tau) - g(\xs(\tau),\tau)||_\Sigmab^2  \right]
\end{equation}
where $q$ is the distribution over the continuous in time stochastic process. The expectation can be estimated with Monte Carlo as,
\begin{equation}
\KL(q||p) \approx \KL(q_0 || p_0) + \frac{1}{M}  \sum_{m=1}^M  \int_0^{t_f} dt \left[ ||f(\xs_m(t),t) - g(\xs_m(t),t)||_\Sigma^2  \right]
\end{equation}
where $\xs_m$ are sample trajectories.

Theorem 30.12.4 of \cref{Lapidoth_2017} provides the inequality,
\begin{equation}
\left( \int d\xs_{[0,n-1]} q_{[0,n]} \right) \left[ \ln \frac{ \int d\xs_{[0,n-1]} q_{[0,n]} } { \int d\xs_{[0,n-1]} p_{[0,n]} } \right]
\leq \int d\xs_{[0,n-1]} \left[ q_{[0,n]} \ln \frac{ q_{[0,n]} } { p_{[0,n]} } \right]
\end{equation}
Integrating both sides we find,
\begin{equation}
\begin{aligned}
\int d\xs_n \,\left( \int d\xs_{[0,n-1]} q_{[0,n]} \right) \left[ \ln \frac{ \int d\xs_{[0,n-1]} q_{[0,n]} } { \int d\xs_{[0,n-1]} p_{[0,n]} } \right]
\leq \int d\xs_n \, \int d\xs_{[0,n-1]} \left[ q_{[0,n]} \ln \frac{ q_{[0,n]} } { p_{[0,n]} } \right] \\
\int d\xs_n \, \left( \int d\xs_{[0,n-1]} q_{[0,n]} \right) \left[ \ln \frac{ \int d\xs_{[0,n-1]} q_{[0,n]} } { \int d\xs_{[0,n-1]} p_{[0,n]} } \right]
\leq \int d\xs_{[0,n]} \, \left[ q_{[0,n]} \ln \frac{ q_{[0,n]} } { p_{[0,n]} } \right] \\
\int d\xs_n q_n \ln \frac{ q_n } { p_n }
\leq \int d\xs_{[0,n]} \, \left[ q_{[0,n]} \ln \frac{ q_{[0,n]} } { p_{[0,n]} } \right] \\
\KL(q_n||p_n)  \leq \KL(q_{[0,n]}||p_{[0,n]})
\end{aligned}
\end{equation}
which provides a bound on the KL divergence of states at corresponding steps.

\setcounter{equation}{0}
\renewcommand{\theequation}{B-\arabic{equation}}
\section{Langevin dynamics} \label{app:OU}

Generalized Langevin dynamics are described by the SDE
\begin{equation}
\dd\ys = \fs(\ys,t)\dt + \etab
\end{equation}
where $\etab$ is a noise process.
For this work we only need to examine classical, linear Langevin dynamics given by
\begin{equation}
\dd\ys = -\Gammab ({\ys} - \bar{\ys})\dt + \sqrt{\Sigmab} \dd\Bs_t
\end{equation}
where $\Gammab$ is a positive definite matrix, $\Sigmab$ is a covariance matrix, and $\dd\Bs_t$ is a standard multidimensional Brownian motion.
Diagonalizing $\Sigmab$ leads to a set of uncorrelated 1D Langevin dynamics processes.

The classical 1D Langevin equation is an SDE
\begin{equation} \label{eq:OU_sde}
\dd v = \underbrace{-\gamma ({v} - \bar{v})}_{f(v)} \dd t + \underbrace{\sigma \dd B_t}_{\eta(t)}
\end{equation}
where $\gamma$ and $\sigma$ are parameters controlling the decay to the mean $\bar{v}$ and the magnitude of the random force $\eta$.
It is also known as the Ornstein-Uhlenbeck (OU) process.
\eref{eq:OU_sde} has the solution
\begin{equation}
v(t) = \bar{v} + (v_0-\bar{v}) \exp\left( -\gamma t \right)
+ \int_0^t \mathrm{d}s \, \exp\left( -\gamma(t-s) \right)
\eta(s)
\end{equation}
where the integral is an It\^{o} integral.
The mean $\langle v \rangle \to \bar{v}$ converges to $\bar{v}$ since $\langle \dd B_\tau \rangle = 0$
while the variance approaches
$\langle (v-\bar{v})^2 \rangle
\to \frac{1}{\gamma} \langle \eta^2 \rangle$
since
\begin{equation} \label{eq:FD}
\langle (v-\bar{v})^2 \rangle
= \lim_{t\to\infty} \left[
v_0^2 \exp\left( -2 \gamma t \right)
+  \frac{\sigma^2}{\gamma} \left( 1 - \exp\left( -2 \gamma t \right) \right)
\right]
= \frac{\sigma^2}{\gamma}
= \frac{1}{\gamma} \langle \eta^2 \rangle
\end{equation}
at the statistical steady state.
\eref{eq:FD} is a form of the fluctuation-dissipation theorem stating that the variance of the solution is determined by the balance between the drag/drift and the random force/excitation.

Clearly the energy $U(v)$ associated with the drift is
\begin{equation}
U(v) = \frac{\gamma}{2} (v-\bar{v})^2
\end{equation}
so that
\begin{equation}
\nabla_v U(v)
= -f(v)
=  \gamma (v-\bar{v}) \ .
\end{equation}
The Langevin equation can be reformulated as a Fokker–Planck equation that governs the probability distribution of the random variable which results in a normal state state distribution
\begin{equation}
\prob_\infty(v) = \sqrt{\frac{\gamma}{\pi \sigma^2}} \exp\left(
- \frac{\gamma}{\sigma^2} (v-\bar{v})^2
\right).
\end{equation}
Hence, if $\sigma^2 = 2$ then
\begin{equation}
\log \prob_\infty(v) = U(v) + \text{constant}
\end{equation}
which implies a specific noise magnitude $\sigma = \sqrt{2}$ to sample the distribution connected to the drift.

\setcounter{equation}{0}
\renewcommand{\theequation}{C-\arabic{equation}}
\section{The score function for the Langevin dynamics model} \label{app:LScore}

Here we provide the derivation for the score function corresponding to the parameters of the Langevin dynamics model presented in Section~\ref{sec:res_chem}.
Let $W$ and $\ybr$ be independent Gaussian random variables defined by:
\[
W \sim \mathcal{N}(\mu_W, \sigma_W^2), \quad \ybr \sim \mathcal{N}(\mu_{\ybr}, \sigma_{\ybr}^2).
\]
First, we will derive the joint probability density function (PDF) for $(W, \ybr)$ and then for $(W, b)$ where $b = -W\ybr$.
Given that $W$ and $\ybr$ are independent random variables, their joint PDF is given by
\[
f(W,\ys) = f_W(W) \cdot f_{\ys}(\ys) =  \frac{1}{\sqrt{2\pi \sigma_W^2}} \frac{1}{\sqrt{2\pi \sigma_{\ybr}^2}}
\exp\left({-\frac{(W - \mu_W)^2}{2\sigma_W^2}} \right)
\exp\left({-\frac{(\ybr - \mu_{\ybr})^2}{2\sigma_{\ybr}^2}}\right) .
\]
To get to the PDF of $(W,b)$, we first set $\ybr$ to  $\ybr =  -b/W$. The Jacobian of the transformation from $(W,\ybr)$ to $(W, b)$ is:
\[
J = \begin{vmatrix}
\frac{\partial W}{\partial W} & \frac{\partial W}{\partial \ybr} \\
\frac{\partial b}{\partial W} & \frac{\partial b}{\partial \ybr}
\end{vmatrix} = \begin{vmatrix}
1 & 0 \\
-\ybr & -W
\end{vmatrix} = -W.
\]
The joint PDF of $(W,b)$ is then given by:
\[
f(W,b) = f(W, -\frac{b}{W}) \cdot \frac{1}{\vert J\vert}
= \frac{1}{|W| 2\pi \sigma_W\sigma_{\ybr}}
\exp\left({-\frac{(W - \mu_W)^2}{2\sigma_W^2}}\right)
\exp\left({-\frac{\left(b/W + \mu_{\ybr}\right)^2}{2\sigma_{\ybr}^2}}\right).
\]
The logarithm of the joint PDF:
\begin{equation}
\log f(W,b) = -\log |W| - \log(2\pi)-\log \sigma_W -  \log \sigma_{\ybr} - \frac{(W - \mu_W)^2}{2\sigma_W^2} - \frac{\left(b/W + \mu_{\ybr}\right)^2}{2\sigma_{\ybr}^2}.
\end{equation}
Finally, the partial derivatives w.r.t $W$ and $b$ are given by:
\begin{eqnarray}
\frac{\partial \log f(W,b)}{\partial W} &=& -\frac{1}{W} - \frac{W - \mu_W}{\sigma_W^2} + \frac{b}{\sigma_{\ybr}^2 W^2} \left(b/W + \mu_{\ybr}\right)\\
\frac{\partial \log f(W,b)}{\partial b} &=& -\frac{1}{\sigma_{\ybr}^2 W} \left(b/W + \mu_{\ybr}\right).
\end{eqnarray}

\end{document}